\documentclass[runningheads]{llncs}

\usepackage{eccv}

\usepackage{eccvabbrv}

\usepackage{graphicx}
\usepackage{booktabs}

\usepackage[accsupp]{axessibility}  

\usepackage[pagebackref,breaklinks,colorlinks,citecolor=eccvblue]{hyperref}

\usepackage{orcidlink}

\usepackage{amsmath}
\usepackage{url}
\usepackage{multirow}
\usepackage{siunitx}
\usepackage{graphbox}
\usepackage{hhline}
\usepackage{subcaption}
\usepackage{tikz}
\usepackage{graphicx}
\usetikzlibrary{positioning}

\DeclareMathAlphabet\mathbfcal{OMS}{cmsy}{b}{n}

\newcommand{\source}{\alpha}
\newcommand{\target}{\beta}

\newcommand{\distribution}{\mathcal{D}}
\newcommand{\labelset}{\mathbb{Y}}
\newcommand{\imageset}{\mathbb{I}}
\newcommand{\heatmapset}{\mathbb{H}}
\newcommand{\pseudogtset}{\mathbb{P}}

\makeatletter
\DeclareRobustCommand\onedot{\futurelet\@let@token\@onedot}
\def\@onedot{\ifx\@let@token.\else.\null\fi\xspace}
\def\eg{\emph{e.g}\onedot} 
\def\ie{\emph{i.e}\onedot} 
 
 \def\vs{\emph{vs}\onedot}

\makeatother

\DeclareMathOperator*{\argmax}{arg\,max}
\DeclareMathOperator*{\argmin}{arg\,min}

\newcommand{\definemodel}[1]{\textbf{#1}}
\newcommand{\StudNoSNoOF}[1]{\textit{St-M-OF}}
\newcommand{\StudSNoOF}[1]{\textit{St+M-OF}}
\newcommand{\StudSOF}[1]{\textit{St+M+OF}}

\newcommand{\newtext}[1]{\textcolor{black}{#1}}
\newcommand{\reviewersugg}[1]{\textcolor{black}{#1}}

\usepackage{enumitem}

\begin{document}

\title{Adapting Fine-Grained Cross-View Localization to Areas without Fine Ground Truth} 

\titlerunning{Adapting Cross-View Localization without Fine Ground Truth}

\author{Zimin Xia\inst{1}\orcidlink{0000−0002−4981−9514} \and
Yujiao Shi\inst{2}\orcidlink{0000-0001-6028-9051} \and
Hongdong Li\inst{3}\orcidlink{0000-0003-4125-1554} \and
Julian F. P. Kooij\inst{4}\orcidlink{0000−0001−9919−0710}}

\authorrunning{Z.~Xia et al.}


\institute{École Polytechnique Fédérale de Lausanne (EPFL), Switzerland
\email{zimin.xia@epfl.ch}\\
\and
ShanghaiTech University, China\\
\and
Australian National University, Australia \\
\and
Delft University of Technology, The Netherlands}

\maketitle

\begin{abstract}
Given a ground-level query image and a geo-referenced aerial image that covers the query's local surroundings, fine-grained cross-view localization aims to estimate the location of the ground camera inside the aerial image.
Recent works have focused on developing advanced networks trained with \newtext{accurate} ground truth (GT) locations of ground images. 
However, the trained models always suffer a performance drop when applied to images in a new target area that differs from training. 
In most deployment scenarios, acquiring \newtext{fine} GT, \newtext{\ie accurate GT locations,} for target-area images to re-train the network can be expensive and sometimes infeasible. 
In contrast, collecting images with \reviewersugg{noisy} GT with errors of tens of meters is often easy. 
Motivated by this, our paper focuses on improving the \newtext{performance} of a trained model \newtext{ in a new target area} by leveraging only the target-area images without \newtext{fine} GT. 
We propose a weakly supervised learning approach based on knowledge self-distillation. This approach uses predictions from a pre-trained model as pseudo GT to supervise a copy of itself.
Our approach includes a mode-based pseudo GT generation for reducing uncertainty in pseudo GT and an outlier filtering method to remove unreliable pseudo GT. 
Our approach is validated using two recent state-of-the-art models on two benchmarks. 
The results demonstrate that it consistently and considerably boosts the localization accuracy in the target area. 
\end{abstract}

\section{Introduction}

\begin{figure}
    \centering
    \includegraphics[width=1\linewidth]{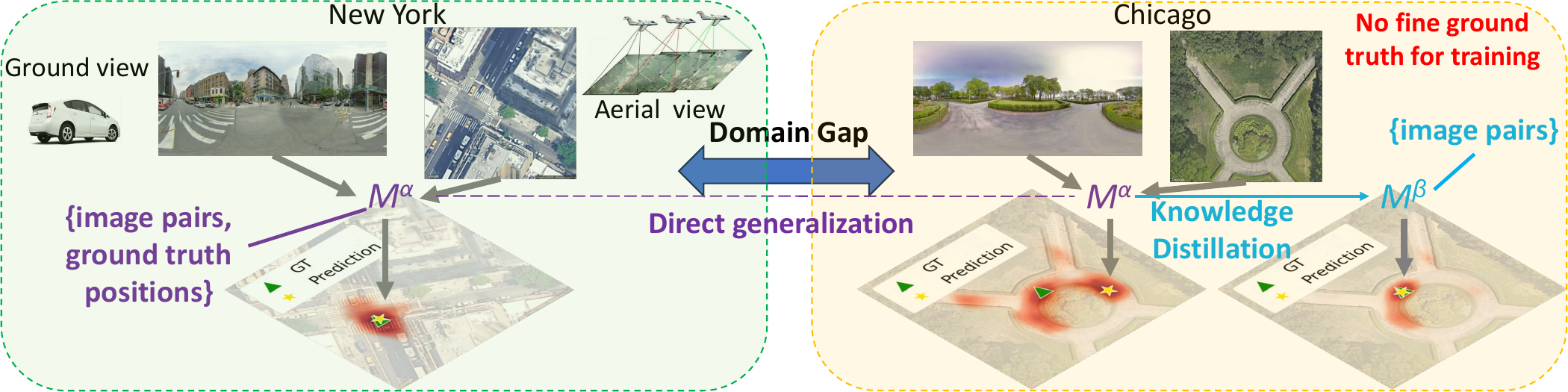}
    \captionof{figure}{
    Learning-based cross-view localization models often perform well when test images are from the same area used in training, as shown in the green box.
    When inference in a new target area where no \newtext{fine} ground truth is available, the standard practice (in purple) directly deploys a model trained in a different area, leaving an obvious domain gap. 
    Due to this domain gap, the direct generalization often results in a performance drop, causing uncertain or erroneous predictions.
    To address this, we propose a knowledge self-distillation-based weakly-supervised learning approach (in cyan) to adapt the model to the target area using only ground-aerial image pairs without \newtext{fine} ground truth locations. This leads to better localization performance.
    }
    \label{fig:figure1}
\end{figure}

Visual localization, a fundamental task in vision and mobile robotics, aims to identify the location of a camera only from the images it takes. Commonly, the image is compared to a pre-constructed map.
However, constructing a suitable map with traditional survey-grade mapping vehicles (often equipped with cameras, LiDAR, and high-precision GNSS sensors) 
is both laborious and expensive. 
On the other hand, aerial or satellite images provide global coverage and become more easily accessible, making them promising map sources.
In this work, we focus on the task of fine-grained cross-view localization to pinpoint the precise geospatial location of a ground camera within a geo-referenced aerial image patch covering local surroundings.
\newtext{The key underlying assumption of this task~\cite{xia2022visual,shi2022beyond,lentsch2023slicematch,fervers2023uncertainty,xia2023convolutional,shi2023boosting,wang2023fine} is that although we \textit{do not} have an accurate fine-grained location of the ground camera, we \textit{do} have a \reviewersugg{\textit{noisy}} localization prior available at inference time to identify the aerial image that covers the ground camera's location.}
For applications such as autonomous driving, fine-grained cross-view localization is a viable supplement to traditional positioning sensors, such as GNSS, especially in urban canyons where the GNSS positioning error can reach tens of meters~\cite{benmoshe2011urbangnss}.

%

As shown in Figure~\ref{fig:figure1}, there are two main scenarios in cross-view localization.
\textbf{(1)}~Same-area testing (Figure~\ref{fig:figure1}, green box): When the \newtext{fine ground truth, \ie the accurate location of the ground camera,} is available in the target area, a cross-view localization model can be trained on this data and then deployed for inference on new test images.
\textbf{(2)}~Cross-area testing (Figure~\ref{fig:figure1}, yellow box, left): When there is no \newtext{fine} ground truth in the target area, it is common to train the model on images from a different area for which \newtext{fine} ground truth is available, and then the trained model is directly deployed in the target area.
Because of the domain gap between the two areas, the predicted location becomes less reliable.
Although many works~\cite{zhu2021vigor,xia2022visual,shi2022beyond,lentsch2023slicematch,fervers2023uncertainty,xia2023convolutional,shi2023boosting} have been proposed for \newtext{fine-grained} cross-view localization, they all suffer from this performance drop when directly deploying in a new target area.
\newtext{Nevertheless}, this cross-area scenario is more realistic for real-world use cases, since collecting \newtext{fine ground truth} of every region is expensive and sometimes infeasible.
\newtext{Recent works~\cite{shi2022beyond,fervers2023uncertainty,lentsch2023slicematch} even found errors in ground truth locations in popular datasets~\cite{Geiger2013IJRR,agarwal2020ford,wilson2023argoverse,zhu2021vigor}.}
\newtext{Therefore, an alternative to fully-supervised training on fine ground truth is needed to scale cross-view localization models to larger areas.}

\newtext{We propose to address this problem of cross-area localization by relying on the exact same key assumption in the fine-grained cross-view localization task.
}
\newtext{Namely,} it is straightforward to collect ground images with \newtext{\reviewersugg{noisy} ground truth, \ie the \reviewersugg{rough} location of the ground camera, at a new area to 
identify the local aerial image patch.}
\newtext{For instance, inaccurate GNSS measurements in urban canyons are unreliable as fine ground truth~\cite{benmoshe2011urbangnss}, but can still be used as \reviewersugg{noisy} localization prior.}
\newtext{Then}, our goal is to \textit{improve a pre-trained model's localization performance in the target area by leveraging only the ground-aerial image pairs in the target area, without associated \newtext{fine} ground truth locations\footnote{Recent models need the ground camera's orientation for training.  
We assume the camera orientation is known since it can be acquired easily, \eg by \newtext{the} digital compass in \newtext{a mobile phone or a vehicle.}}.}


For \newtext{this goal}, we adopt knowledge self-distillation~\cite{furlanello2018born,wang2021knowledge} to finetune a fine-grained cross-view localization model in a weakly-supervised manner in which only \reviewersugg{rough} location is used for pairing the ground and aerial images.
We use a model pre-trained from another area as the teacher model to generate pseudo ground truth for the target-area images and use it to train a student model, which is initialized as a copy of the teacher model.
Since the teacher's output can be uncertain in the target area, directly using it as pseudo ground truth might reinforce incorrect localization estimates and lead to sub-optimal results.
\newtext{We address this by introducing methods to reduce the uncertainty and filter out the outliers in the pseudo ground truth.}
\newtext{Concretely,} our contributions are\footnote{Our code will be released to facilitate reproducible research.}:

\textbf{(1)} We propose a knowledge self-distillation-based weakly-supervised learning approach that considerably improves models' localization performance in a new area by only leveraging the ground-aerial image pairs without ground truth locations.
The proposed approach is validated using two state-of-the-art methods on two benchmarks.
\textbf{(2)} For methods with coarse-to-fine outputs,
we investigate how to reduce the uncertainty and suppress the noise in teacher model's predictions.
Using our proposed single-modal pseudo ground truth leads to a better student model than using the multi-modal heat maps from the teacher model.
\textbf{(3)} We design a simple but effective method for filtering outliers in the pseudo ground truth. Training with filtered pseudo ground truth further improves the localization accuracy of the student model.


\section{Related Work}

\textbf{Cross-view localization} is formulated differently depending on the use case.
For large-scale coarse localization, a common formulation is image retrieval~\cite{ground-to-aerialgeolocalization, WideAreaImageGeolocalization,hu2018cvm,shi2019spatial,liu2019lending,regmi2019bridging,yang2021cross,zhu2022transgeo,shi2022accurate,toker2021coming}.
In this setting, the continuous aerial imagery is divided into small patches.
The ground query image's location is approximated by the retrieved patch's geolocation.
However, for fine-grained localization, image retrieval methods need to sample the patch densely~\cite{xia2020geographically,9449965}, and it increases both computation and storage usage.

Recently, there have been increasing attempts to estimate the precise location directly, sometimes together with the orientation, of the ground camera on a known aerial image patch.
In \cite{zhu2021vigor}, the location offset between the ground query and the aerial image is regressed based on their image descriptors.
Instead of regression, \cite{xia2022visual} formulated the localization task as a dense classification problem to capture the multi-modal localization uncertainty.
Later, this idea is extended by~\cite{xia2023convolutional} to include coarse-to-fine predictions and build orientation equivariant ground image descriptors.
Several works~\cite{shi2022beyond,shi2022cvlnet,wang2023fine} explored the geometry transformation between ground and aerial views.
\cite{shi2022beyond}~estimated the ground camera pose using the iterative Levenberg–Marquardt algorithm and \cite{wang2023fine} made use of a deep homography estimator~\cite{cao2022iterative} to infer the ground camera pose.
In \cite{fervers2023uncertainty,sarlin2023orienternet,sarlin2023snap,shi2023boosting}, the ground camera pose is estimated by densely comparing a Bird's Eye View (BEV) representation constructed using ground images to an aerial representation. 
SliceMatch~\cite{lentsch2023slicematch} took an efficient generative testing approach to select the most probable pose from a candidate set.
Commonly, the localization output is represented as a heat map~\cite{xia2022visual,xia2023convolutional,fervers2023uncertainty,lentsch2023slicematch,shi2023boosting,wang2023fine}, where the value at each location (\ie pixel in the aerial image) denotes how likely the ground camera locates there, and state-of-the-arts~\cite{xia2023convolutional,shi2023boosting} construct the heat map in a coarse-to-fine manner.
Despite extensive methodological consideration, the performance of the above approaches dropped considerably
when directly generalizing to images collected in an area that differs from the training set. 
In this work, we aim to bridge this gap.

\textbf{Unsupervised domain adaptation} (UDA) is a well-studied problem in many other vision tasks~\cite{zhang2021survey,wang2018deep}.
\newtext{The objective is to adapt a model trained in the source domain to the target domain without labels from the target domain, such that the adapted model can perform well on the test samples from the target domain.}
\reviewersugg{More specifically, UDA can be categorized as source-free~\cite{jing2023order,zheng2021rectifying,lu2023uncertainty,litrico2023guiding} and non-source-free~\cite{ji2021refine,cardace2023self,xia2023transferable,tarvainen2017mean,zhou2022uncertainty,wang2021uncertainty,guan2021uncertainty,lai2023padclip} depending on if the source domain labels are used during adaptation.}
\reviewersugg{To minimize the discrepancy between features from the source and target domain, some works~\cite{long2015learning,sun2016deep,gholami2020unsupervised} use manually crafted metrics to measure this discrepancy.
Adversarial methods~\cite{tzeng2017adversarial,long2018conditional,zhang2019domain} deploy a discriminator to achieve this.}
\reviewersugg{\cite{sohn2020fixmatch,berthelot2019mixmatch,vu2019advent} observed that predictions in the target domain often contain more uncertainty than those in the source domain.}
Hence, additional objectives, e.g. entropy minimization~\cite{grandvalet2004semi}, are included for training the model in a semi-supervised manner using images from both the source and target domain.
Another promising type of domain adaptation is based on pseudo labels~\cite{xie2018learning,zhang2018collaborative}.
It bears similarities to \textbf{knowledge distillation} (KD)~\cite{buciluǎ2006model}.
KD's primary objective is to transmit the knowledge acquired by a more comprehensive teacher model to a smaller student model~\cite{wang2021knowledge,gou2021knowledge}. 
Knowledge self-distillation, in which the teacher and student share the same architecture, is a special branch of KD pioneered by Born-Again Networks~\cite{furlanello2018born}.
The key idea is to use the model from the previous step to generate pseudo labels for training the model at the current step.
Recent works~\cite{ding2022kd,zhang2019your,hou2019learning,ji2021refine,zhang2021self,an2022efficient,song2023multi} also tried to use the information from deeper layers to supervise the shallower layers inside the model.
\newtext{To apply KD for UDA}, the teacher model generates pseudo labels in the target domain to adapt the student model~\cite{nguyen2021unsupervised,feng2021kd3a,zhou2021domain}.
\reviewersugg{Since the pseudo labels are not always reliable, uncertain ones, e.g. measured by their entropy~\cite{zhou2022uncertainty,wang2021uncertainty,litrico2023guiding}, are often filtered in student learning.}
\reviewersugg{However, such measures are developed for purely categorical tasks.
The classes in localization heat maps are spatially ordered.}

\section{Methodology}
\reviewersugg{The most desirable real-world setup is to adapt a pre-trained model to the target area without requiring access to (perhaps licensed or high-volume) source-domain data. 
Our scope is thus source-free UDA.
We first formalize the fine-grained cross-view localization task.
Then, we introduce our proposed approach.}

\subsection{Task Definition}

Given a ground-level image $G$ and an aerial image $A$ that covers the local surroundings of $G$, the task of fine-grained cross-view localization is to determine the image coordinates $\hat{y}=(\hat{u},\hat{v})$ of the ground camera within \newtext{aerial image} $A$, where $\hat{u} \in [0,1]$ and $\hat{v} \in [0,1]$.
Recent methods~\cite{xia2022visual,xia2023convolutional,fervers2023uncertainty,lentsch2023slicematch,shi2023boosting} achieve this task by training a deep model $\mathcal{M}(G,A)$
which predicts a \textit{heat map} $H$ to capture the underlying localization confidence over spatial locations,
and the most confident location can be used as predicted location $y$,
\begin{align}
    H = \mathcal{M}(G, A), \quad y = \argmax_{u,v} (H(u,v)).
    \label{eq:heatmap_prediction}
\end{align}

To optimize the model's parameters $\theta_\source$
with respect to a model specific loss functions $\mathcal{L}_{\mathcal{M}}$,
an annotated dataset of a set of $N_\source$ ground-aerial image pairs, $\imageset_\source=\{ \{G_1,A_1\}, ..., \{G_{N_\source},A_{N_\source}\}\}$, and their corresponding \newtext{fine}  ground truth $\labelset_\source = \{ \hat{y}_1, ..., \hat{y}_{N_\source}\}$ is used,
\begin{align}
    \theta_\source = \argmin_{\theta} \mathbb{E}_{\{G, A\} \in \imageset_\source, \hat{y} \in \labelset_\source} \left[ \mathcal{L}_{\mathcal{M}}( \mathcal{M}(G, A \mid \theta), \hat{y}) \right].
    \label{eq:supervised_learning}
\end{align}
%
%
The training image set $\imageset_\source$ consists of samples drawn from a true distribution $\distribution_\source$ representing a specific geographic area~$\source$, \ie $\imageset_\source \overset{\mathrm{i.i.d.}}{\sim} \distribution_\source$.
When the model is deployed, the test image set $\imageset_{test}$ can either come from the \textit{same area} $\source$, or a new environment $\target$.
As motivated before, we focus on the \textit{cross-area} setting, namely $\imageset_{test}$ is from the target area $\target$, \ie $\imageset_{test} \overset{\mathrm{i.i.d.}}{\sim} \distribution_\target$.
Because of the domain gap, $\distribution_\target \neq \distribution_\source$, directly deploying the trained model $\mathcal{M}^{\source} := \mathcal{M}(\cdot \mid \theta^{\source})$ on test set $\imageset_{test}$ as in current practice is sub-optimal.

It is important to note that \newtext{standard fine-grained cross-view localization~\cite{xia2023convolutional,fervers2023uncertainty,lentsch2023slicematch,shi2023boosting} assumes the pairing between ground and aerial images is known during inference, as} collecting ground-level images with \reviewersugg{rough} location estimates in the target area is often easy.
Therefore, \newtext{we propose to consider the easily available pairing information for weakly-supervised learning by collecting} another set of images $\imageset_\target=\{ \{G_1,A_1\}, ..., \{G_{N_\target},A_{N_\target}\}\}$ from the target area~$\target$, $\imageset_\target \overset{\mathrm{i.i.d.}}{\sim} \distribution_\target$, without corresponding \newtext{fine ground truth} $\labelset_\target$.
As noted before, the orientation of the ground camera is assumed known.

Our objective is then to adapt a fine-grained cross-view localization model $\mathcal{M}^{\source}$ to the target area $\target$ by leveraging the image set $\imageset_\target$ without \newtext{fine} ground truth locations such that the model performance on $\imageset_{test}$ can be improved.

\subsection{UDA for Cross-View Localization}

So far, no prior work addressed the task of adapting fine-grained cross-view localization to new areas without \newtext{fine ground truth}.
To decide on a suitable UDA approach, we first note that
heat maps of state-of-the-art models reflect more uncertainty for cross-area samples than for same-area samples~\cite{lentsch2023slicematch,xia2023convolutional,shi2023boosting}.
The higher uncertainty results in more small positional errors, but also more modes in the heat map, yielding more outliers with large positional errors.

We therefore consider in this work UDA techniques that can help reduce the uncertainty.
One option is \textit{entropy minimization}~\cite{grandvalet2004semi},
namely
to directly deploy the trained model $\mathcal{M}^{\source}$ on the image set $\imageset_\target$ and then encourage the final output heat map $H$ to be more certain by minimizing its entropy.
However minimizing the entropy does not necessarily encourage the model to converge towards the correct location for $\{G,A\} \in \imageset_\target$,
as the model may just as well become more confident about the outliers.
Our experiments shall validate entropy minimization's shortcomings for our task.

We instead propose to pursue
\textit{knowledge self-distillation}~\cite{zhang2021self} for our task.
The trained model $\mathcal{M}^\source$ from the source area $\source$ can be used as the teacher model to generate \textit{pseudo ground truth} $X$ for image set $\imageset_\target$ to train a student model $\mathcal{M}^\target$.
Here, we consider $X$ as a target heat map 
with the same spatial resolution as the aerial image $A$.
The student model has the same architecture as the teacher model and is initialized using the teacher model's weights $\theta_\source$.
Encouraging the outputs of the student model to mimic $X$ can improve the accuracy of the student model on images from $\target$,
especially if we control the generation of pseudo ground truth to suppress unwanted modes and select for reliable samples.

Finally, we point out that the recent state-of-the-art methods~\cite{xia2023convolutional,shi2023boosting} have $K$ coarse-to-fine heat map outputs, \ie $\heatmapset = \mathcal{M}(G, A)$ and $\heatmapset = \{H_1, ..., H_K\}$.
The spatial resolution of the next level heat map is higher than that of the previous level, namely $\text{res}(H_{k+1}) > \text{res}(H_k)$ where $k$ is the index for the level and $\text{res}()$ returns the spatial resolution.
The final predicted location then becomes $y = \argmax_{u,v} (H_K(u,v))$.
For other applications with coarse-to-fine models, encouraging shallower layers' activation to mimic deeper layers' activation can bootstrap model performance~\cite{zhang2021self}.
Similarly, knowledge self-distillation for cross-view localization may also exploit such coarse-to-fine maps.

\subsection{Proposed Approach}

\begin{figure*}[t]
    \centering
    \includegraphics[width=1\linewidth]{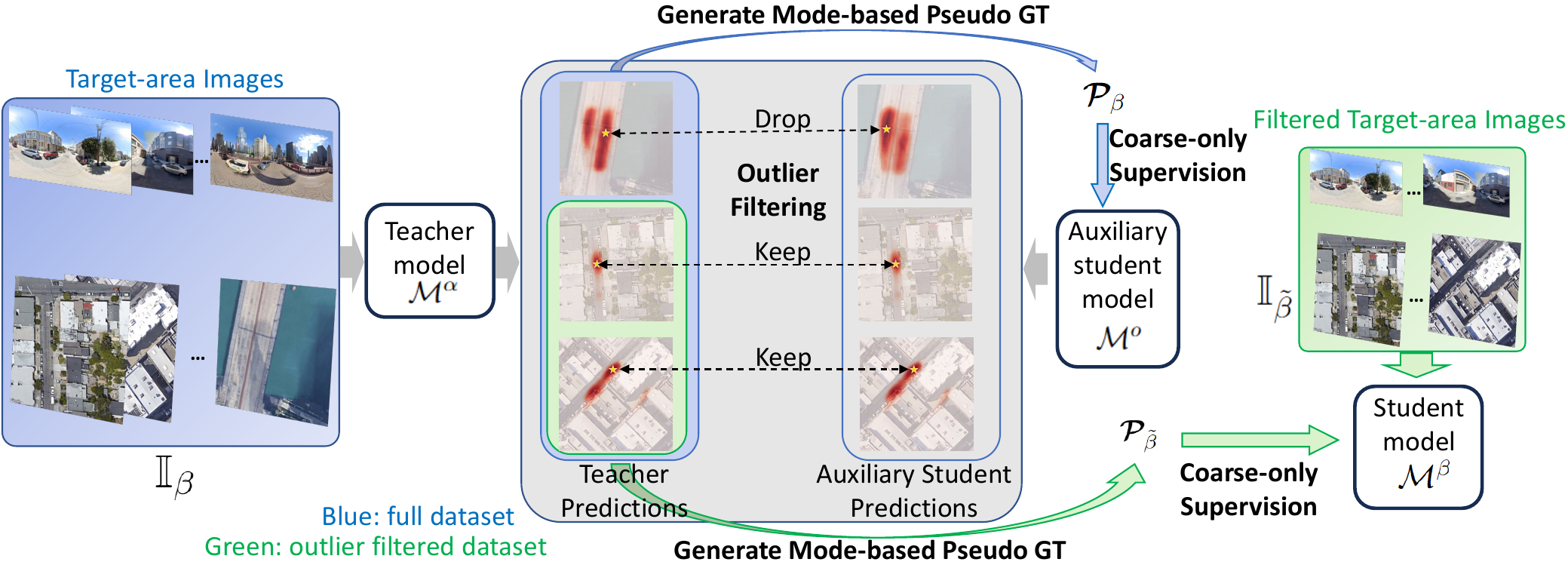}
    \caption{Overview of our proposed weakly-supervised learning approach. 
    We first employ a teacher model trained on data from another area to generate pseudo GT, $\mathbfcal{P}_\target$, on target-area images, shown in blue. The pseudo GT is then used to train an auxiliary student model $\mathcal{M}_o$.
    After that, we compare the predictions from the teacher model and those from the auxiliary student model, and filter out unreliable teacher predictions (the middle grey box of this figure). The remaining predictions with their pseudo GT, $\mathbfcal{P}_{\Tilde{\target}}$, are used to train our final student model $\mathcal{M}_\beta$, shown in green. 
    }
    \label{fig:KD_pipeline}
\end{figure*}

Usually, the deeper layers in the model have access to more information than the shallower layers, \eg the fine-grained scene layout information passed by the skipped connections, as in UNet~\cite{ronneberger2015u}.
Hence, the output from deeper layers can be more precise than that from shallower layers.
We therefore propose to follow the
``Best Teacher Distillation'' paradigm~\cite{zhang2021self} and generate pseudo ground truth $X$ from only the highest-resolution heat map predicted by the teacher model on the target domain input.

A naive approach is, for any $\{G,A\} \in \imageset_\target$, using simply $X := H_K^{\source}$
from teacher output\footnote{Note that we use superscript $\source$ to indicate output from model $\mathcal{M}^\source$.}
$\{ H_1^{\source}, \cdots, H_K^{\source} \} = \mathcal{M}^\source(G, A)$.
Then, this high-resolution pseudo ground truth $X$ is down-sampled to create a set of pseudo ground truth heat maps $\pseudogtset = \{P_1, ..., P_K\}$ to supervise the student model at all levels,
\begin{align}
    P_k = \text{downsample}_k(X) \quad \text{s.t.} \quad \text{res}(P_k) = \text{res}(H_k).\label{eq:pseudo_ground_truth}
\end{align}
The set $\mathbfcal{P}_\target = \{\pseudogtset_1, ..., \pseudogtset_{N_\target}\}$ is the complete pseudo ground truth for image set $\imageset_\target$ in the target area for training the student model, where $N_\target$ is the number of the ground-aerial image pairs in $\imageset_\target$.

However, since the pseudo ground truth $X$ contains errors, directly following this naive approach might propagate the errors to the student model $\mathcal{M}^\target$.
Thus, we present several strategies to reduce the teacher's uncertainty, and deal with noise and large outliers in $X$.
Our proposed designs are highlighted \newtext{in bold} in the overview of the approach in Figure~\ref{fig:KD_pipeline}.


\textbf{Coarse-only Supervision:}
Standard Best Teacher Distillation~\cite{zhang2021self} 
suggests supervising heat maps at all levels of the student model using the pseudo ground truth.
However, 
the spatial accuracy of $X$ is limited, and using $X$ to supervise the high-resolution outputs of the student model might propagate this noise.
We note that the down-sampling in Equation~\ref{eq:pseudo_ground_truth} suppresses such positional noise at the lower resolution $P_k$.
Thus using only the lower level $P_k$ might lead to a better student model.
We therefore consider to only compute the loss 
on student model's outputs $\heatmapset^\target = \mathcal{M}^\target(G, A)$
up to a certain level $K' \leq K$, 
\begin{align}
    \mathcal{L}(\heatmapset^\target, \pseudogtset) = \frac{1}{K'} \sum_{k=1}^{K'} \mathcal{L}_k(H_k^\target, P_k).
\label{eq:total_loss}
\end{align}
Here $K'$ is a hyperparameter,
and $\mathcal{L}_k(H_k^\target, P_k)$ is a weighted sum of infoNCE losses~\cite{oord2018representation},
similar to regular training in~\cite{xia2022visual,xia2023convolutional},
except we use pseudo ground truth $P_k$ as weight,
\begin{align}
    \mathcal{L}_k(H_k^\target, P_k) = \frac{1}{\sum P_k}\sum_{m,n} P^{m,n}_k \cdot \mathcal{L}_{\text{infoNCE}}(H_k^\target \mid (m,n)). \label{eq:loss_at_level_k}
\end{align}
$\mathcal{L}_{\text{infoNCE}}(H_k^\target \mid (m,n))$ denotes an infoNCE loss interpreting $H_k^\target$ as metric learning scores, location $(m,n)$ as the positive class,
and all other locations as the negative class.



\textbf{Mode-based Pseudo Ground Truth:} 
Rather than using $H_K^{\source}$ directly as pseudo ground truth $X$, 
we propose to create a ``clean'' pseudo ground truth $X$ that only represents its mode $y^{\source} = \argmax (H_K^{\source})$.
%
%
We thus provide the student with a training objective that represents less uncertainty for the target domain input than its teacher.
Still, it is common when training fine-grained cross-view localization models,
to apply Gaussian label smoothing~\cite{xia2022visual,fervers2023uncertainty} even with reliable ground truth
to aid the learning objective and increase robustness to remaining errors in the annotation~\cite{muller2019does}.
We similarly apply Gaussian label smoothing centered at $y^{\source}$,
\begin{align}
    X(u,v) = \mathcal{N}((u,v) \mid y^{\source},\, I_2\sigma^{2}), \; \text{res}(X) = \text{res}(A). 
    \label{eq:gaussian_smoothing}
\end{align}
%
In Equation~\ref{eq:gaussian_smoothing}, the standard deviation $\sigma$ is a hyperparameter and $I_2$ is a 2D identity matrix.

\textbf{Outlier Filtering:}
Recent deep learning advances~\cite{oquab2023dinov2} highlighted the importance of using curated data.
Motivated by this principle, we prefer having fewer but more reliable samples of the target domain, over having more samples but with potentially large errors in the pseudo ground truth.
The \textit{Mode-based Pseudo Ground Truth} could force a sample's ground truth to commit to a wrong (outlier) location,
therefore we seek to filter out such samples.
%

We here make another observation:
samples where the predicted locations $y^\alpha$ of a teacher and $y^\beta$ of a student greatly differ, the teacher's predictions were more likely to be outliers compared to 
samples where the teacher and student's predicted locations are more consistent, as we will demonstrate in our experiments.
Thus, we propose to first train another auxiliary student model $\mathcal{M}^o$ on all data from the target domain, and compare its prediction to the teacher's to identify stable predictions with little change in the predicted location.
Then, we only use those reliable non-outlier samples to train the final student model $\mathcal{M}^\target$.
Concretely, we first optimize the auxiliary student model $\mathcal{M}^o$ on all $\imageset_\target$ with $\mathbfcal{P}_\target$ using,
\begin{align}
    \theta_o = \argmin_\theta \mathbb{E}_{\{G, A\} \in \imageset_\target, \pseudogtset \in \mathbfcal{P}_\target} \left[ \mathcal{L}( \mathcal{M}(G, A \mid \theta), \pseudogtset ) \right]. \label{eq:outlier_detection_model}
\end{align}
%
%
%
%
Then, we calculate the L2-distance $d^{\source, o} = \lVert \mathbf{y^{\source} - y^{o}} \rVert_2$ between the image coordinates 
predicted by $\mathcal{M}^\source$ and $\mathcal{M}^o$ to find the potential unreliable $\pseudogtset$.
The resulting distance set $\mathbb{D} = \{d_1^{\source, o}, ..., d_{N_\target}^{\source, o}\}$ is used to keep the top-$T\%$ samples in $\imageset_\target$ that have the smallest $T\%$ distance $d^{\source, o}$.
Denoting the filtered image set as $\imageset_{\Tilde{\target}}$ and corresponding pseudo ground truth as $\mathbfcal{P}_{\Tilde{\target}}$, the final student model $\mathcal{M}^\target$ is optimized using Equation~\ref{eq:outlier_detection_model} by substituting $\imageset_\target$ with $\imageset_{\Tilde{\target}}$ and $\mathbfcal{P}_{\target}$ with $\mathbfcal{P}_{\Tilde{\target}}$. 

\section{Experiments}

We first introduce the two datasets used in this paper and our evaluation metric. 
Then, we discuss two state-of-the-art methods~\cite{xia2023convolutional,shi2023boosting}, based on which the proposed weakly-supervised learning is evaluated, followed by implementation details. 
After this, we provide the test results and a detailed ablation study.

\subsection{Datasets}
We adopt two cross-view localization datasets, VIGOR~\cite{zhu2021vigor} and KITTI~\cite{Geiger2013IJRR}, and focus on their cross-area split.

\textbf{VIGOR} dataset contains ground-level panoramic images and their corresponding aerial images collected in four US cities.
In its cross-area split, the training set contains images from two cities, and the test set is collected from \newtext{two other} cities.
We use the training set to train the teacher model and focus on the cross-area setting in our experiments.
To compare direct generalization and our proposed weakly-supervised learning, we conduct a 70\%, 10\%, and 20\% split on the original cross-area test set to create our weakly-supervised training set (no ground truth locations), validation set, and test set.
We use the validation set for finding the stopping epoch during training, as well as for conducting the ablation study.
Our test set is used for benchmarking our method.
We use the improved VIGOR labels provided by~\cite{lentsch2023slicematch}.

\textbf{KITTI} dataset contains ground-level images with a limited field of view. 
We use the aerial images provided by~\cite{shi2022beyond} and adopt their cross-area setting, where the training and test images are from different areas.
Similar to our settings on the VIGOR dataset, we use the training set to train the teacher model and then split the original cross-area test set into 70\%, 10\%, and 20\% for weakly-supervised training of the student model, validation, and testing.

\subsection{Evaluation Metrics}
We measure the displacement error $\epsilon$ in meters between the predicted location and the ground truth location, \ie $\epsilon = s\lVert y - \hat{y} \rVert_2$, where $s$ is the scaling factor from image coordinates to real-world Euclidean coordinates.
Then, mean and median displacement errors over all samples are reported as our evaluation metrics. 
\newtext{On the KITTI dataset, we further decompose the displacement errors into errors in the longitudinal direction (along the camera's viewing direction, typically along the road), and errors in the lateral direction (perpendicular to the viewing direction), following the common evaluation protocol~\cite{shi2022beyond,lentsch2023slicematch,xia2023convolutional}.}

\subsection{Backbone State-of-the-Art Methods}
Two state-of-the-arts, Convolutional Cross-View Pose Estimation (CCVPE)~\cite{xia2023convolutional} and Geometry-Guided Cross-View Transformer (GGCVT)~\cite{shi2023boosting} are used to test our proposed weakly-supervised learning approach.
Both methods were proposed for fine-grained cross-view localization and orientation estimation, and have a coarse-to-fine architecture.
CCVPE has two separate branches for localization and orientation prediction.
GGCVT uses an orientation estimation block before its location estimator.
In this work, we use them for localization only.
CCVPE has seven levels of heat map outputs, in which the first six heat maps are 3D, with the first two dimensions for localization and the third dimension for orientation. The last heat map is 2D.
GGCVT has three levels of 2D heat map outputs.

\subsection{Implementation Details}

We use the code released by the authors of CCVPE~\cite{xia2023convolutional} and GGCVT~\cite{shi2023boosting} for model implementations.
\newtext{Auxiliary and final student} models are trained following our proposed approach. 
For CCVPE's 3D heat map output, we simply lift the pseudo ground truth heat map $P_k$ to 3D using the known orientation as done in~\cite{xia2023convolutional}.
Following the two model's default settings, we use a batch size of $8$ for CCVPE and $4$ for GGCVT, and a learning rate of $\num{1e-4}$ with Adam optimizer~\cite{kingma2014adam} for both models.

The hyperparameters $K'$, $T$, and $\sigma$ are tuned on the VIGOR validation set.
For CCVPE, we find that including the first two levels of losses, \ie $K'=2$, and $T\% = 80\%$ gives the lowest mean localization error.
For GGCVT, we use all three levels of losses, \ie $K'=3$, and $T\% = 70\%$.
\newtext{We tested $\sigma=1,4,8,12,20$ pixels, and one-hot pseudo ground truth.
Because of $\sigma=4$ gave the best validation result, it is used for both methods.} 
The same setting is directly applied to KITTI. 

\subsection{Results}

We compare the trained student models to teacher models (baselines) on the cross-area test set of VIGOR and KITTI datasets.
Previous state-of-the-art was set by directly deploying CCVPE and GGCVT teacher models to the target area.
On the VIGOR dataset, Table~\ref{tab:test_quantitative} top, the performance of student models trained using proposed weakly-supervised learning surpasses baselines by a large margin.
For CCVPE, our approach reduces the mean and median error by 20\% and 15\% when the orientation of test ground images is unknown.
GGCVT only released its code and models for orientation-aligned \newtext{setting}
for the VIGOR dataset. 
Thus, we follow the same setting.
In this case, our approach reduces 16\% and 5\% mean and median error for GGCVT.
Without extra hyperparameter tuning, we directly use our proposed approach to train models on KITTI, and it again improves the overall localization performance for both models, see Table~\ref{tab:test_quantitative} bottom.

\newtext{We also study the gap between each student model to an Oracle, \ie~the same method using supervised finetuning on fine ground truth at the target area. 
Even though the Oracles still achieve lower errors (CCVPE: Oracle 2.31~m $\vs$ student 3.85~m; GGCVT: Oracle 2.91~m $\vs$ student 4.34~m), we emphasize again that in practice such reliable fine ground truth is generally not available.
Importantly, we also find that when the ground truth does contain errors, using supervised finetuning leads to large test errors, see additional results in our Supplementary Material.
Instead, our weakly-supervised learning approach scales well because it boosts performance at a low cost:
First, there are no extra requirements on the accuracy of localization prior in the target area over previous fine-grained cross-view localization works~\cite{xia2022visual,xia2023convolutional,fervers2023uncertainty,lentsch2023slicematch,shi2023boosting,wang2023fine}, as only ground-aerial image pairs are needed.
Second, since student models are initialized from their teacher, the training time is short.
For example, on VIGOR, using a single 32GB V100 GPU our weakly-supervised learning for CCVPE only adds $\sim6$ hours of training time (including pseudo ground truth generation and outlier filtering) on top of the direct generalization, which has training time of $\sim16$ hours.
}

\begin{table}[ht]
    \caption{Evaluation on VIGOR and KITTI test set. \textbf{Best in bold.} Baseline models are teacher models (previous state-of-the-art). ``Student'' denotes models trained using our proposed weakly-supervised learning without ground truth labels. On VIGOR, we provide test results for both known and unknown orientation cases.
    On KITTI, we test with known orientation.}
    \centering
    \begin{tabular}{p{3.7cm}p{2cm}p{2cm}p{2cm}p{2cm}}
    \toprule
    \multirow{2}{*}{VIGOR, cross-area test} & \multicolumn{2}{c}{Known orientation} & \multicolumn{2}{c}{Unknown orientation} \\
    \cline{2-5}
     & Mean (m) & Median (m) & Mean (m) & Median (m) \\
    \hline
    CCVPE~\cite{xia2023convolutional} & 4.38 & 1.76 & 5.35 & 1.97 \\
    \hline
    CCVPE student (ours) & \textbf{3.85} ($\downarrow 12\%$) & \textbf{1.57} ($\downarrow 11\%$) & \textbf{4.27} ($\downarrow 20\%$) & \textbf{1.67} ($\downarrow 15\%$)\\
    \hhline{=====}
    GGCVT~\cite{shi2023boosting} & 5.19 & 1.39 & - & - \\
    \hline
    GGCVT student (ours)  & \textbf{4.34} ($\downarrow 16\%$) & \textbf{1.32} ($\downarrow 5\%$) & - & - \\
    \midrule
    \midrule
    \multirow{2}{*}{KITTI, cross-area test} & \multicolumn{2}{c}{Longitudinal error} & \multicolumn{2}{c}{Lateral error} \\
    \cline{2-5}
     & Mean (m) & Median (m) & Mean (m) & Median (m) \\
    \hline
    CCVPE~\cite{xia2023convolutional} & 6.55 & 2.55 & 1.82 & \textbf{0.98} \\
    \hline
    CCVPE student (ours) & \textbf{6.18} ($\downarrow 6\%$) & \textbf{2.35} ($\downarrow 8\%$) & \textbf{1.76} ($\downarrow 3\%$) & \textbf{0.98} ($\downarrow 0\%$)\\
    \hhline{=====}
    GGCVT~\cite{shi2023boosting} & 9.27 & 4.66 & 2.19 & 0.85 \\
    \hline
    GGCVT student (ours) & \textbf{8.56} ($\downarrow 8\%$) & \textbf{4.35} ($\downarrow 7\%$) & \textbf{1.90} ($\downarrow 13\%$) & \textbf{0.79} ($\downarrow 7\%$)\\
    \bottomrule
    \end{tabular}
    
    \label{tab:test_quantitative}
\end{table}

\begin{figure}[t]
    \centering
    \begin{minipage}{\linewidth}
    \centering
    \includegraphics[align=c, width=0.33\linewidth]{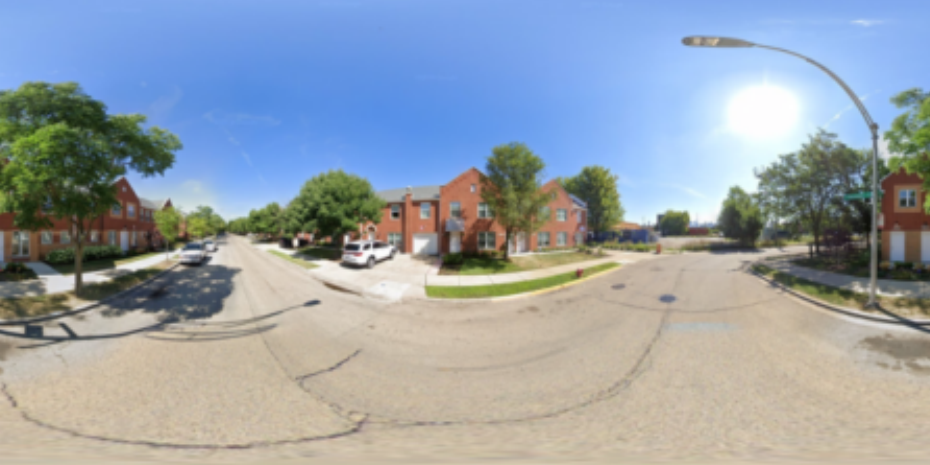}
    \includegraphics[align=c, width=0.3\linewidth]{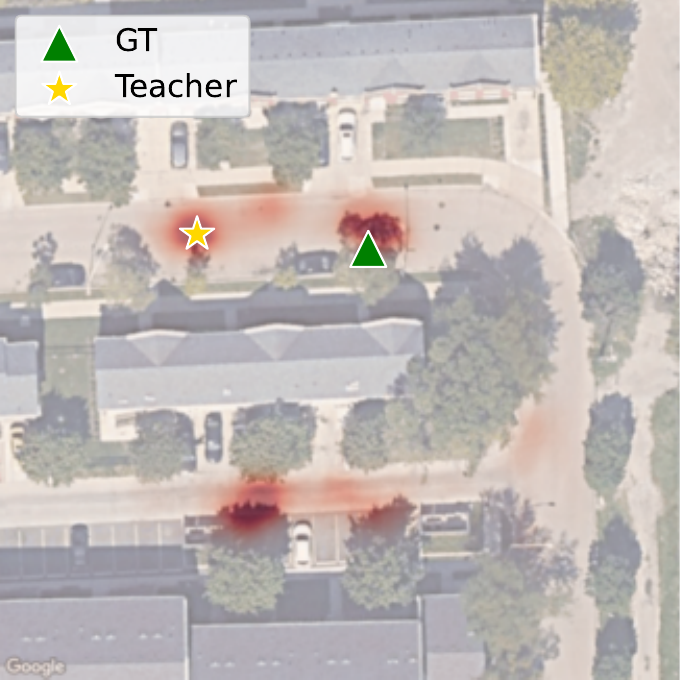}
    \includegraphics[align=c, width=0.3\linewidth]{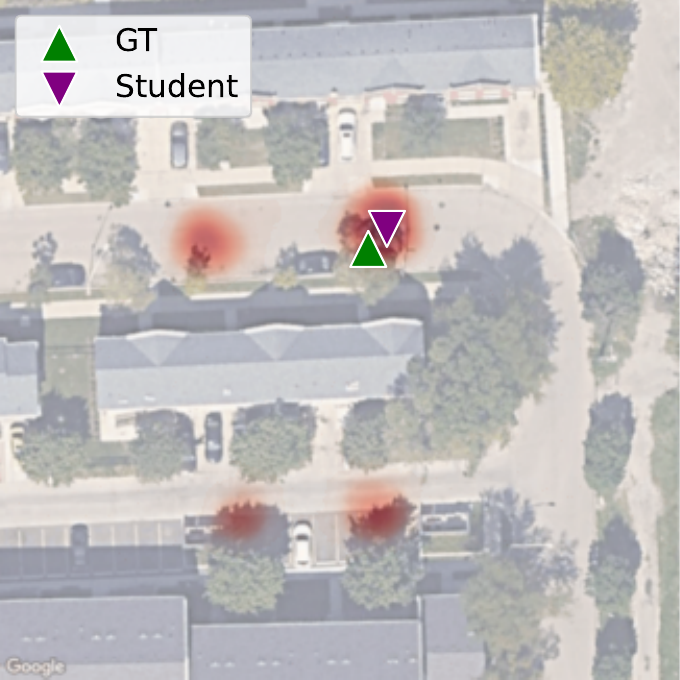} \\
    \includegraphics[align=c, width=0.33\linewidth]{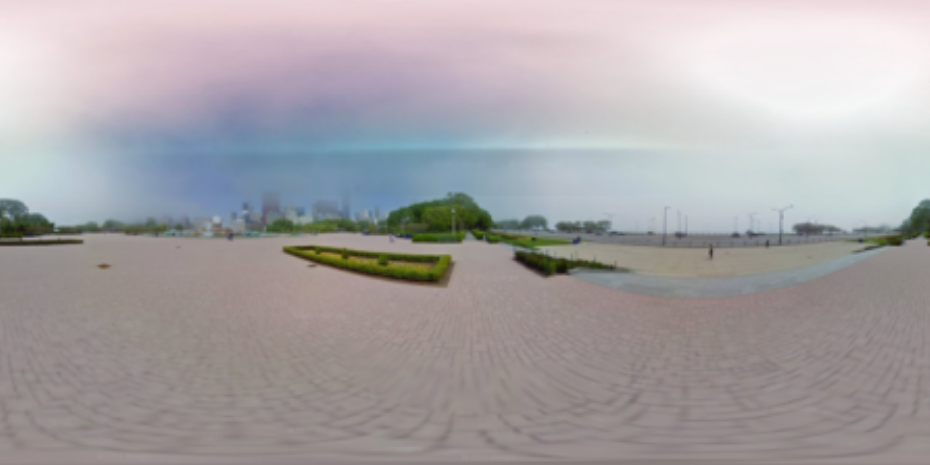}
    \includegraphics[align=c, width=0.3\linewidth]{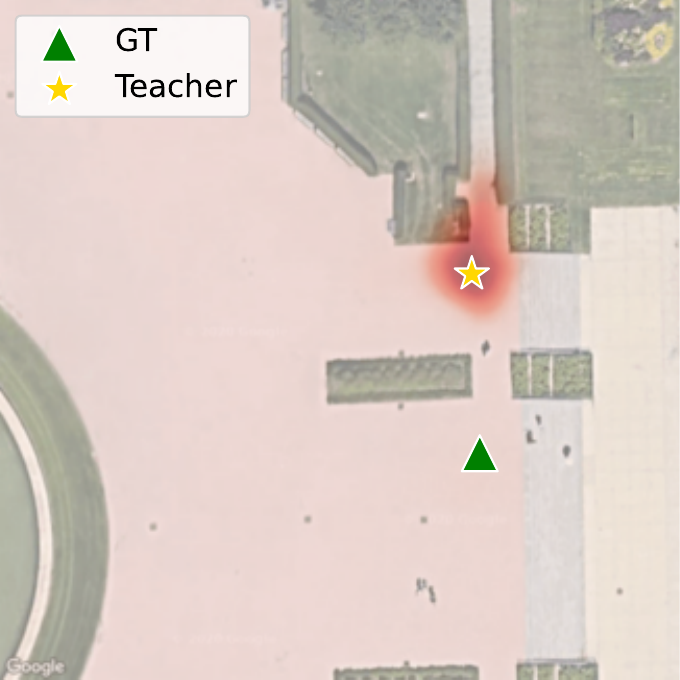}
    \includegraphics[align=c, width=0.3\linewidth]{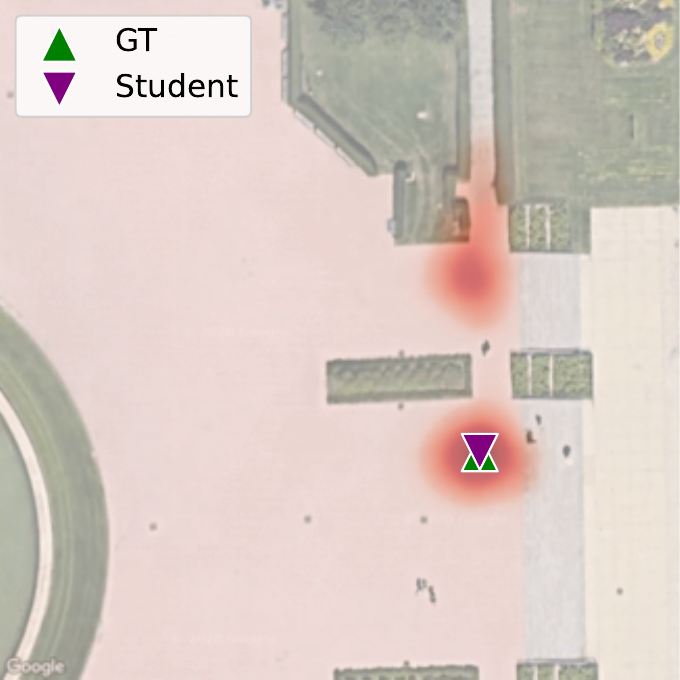}
    \end{minipage}
    \caption{CCVPE teacher and student model's predictions on VIGOR test set. The red color denotes the localization probability (a darker color means a higher probability).}
    \label{fig:CCVPE_qualitative_results}
\end{figure}

Next, we visualize samples where the student model improves over the teacher model.
A typical case is shown in Figure~\ref{fig:CCVPE_qualitative_results} \newtext{top}, in which the teacher model has a multi-modal prediction, and the peak is located in a wrong mode. 
The student model learned to weigh the modes better after adapting to the target environment.
As shown in Figure~\ref{fig:CCVPE_qualitative_results} \newtext{bottom},
sometimes, even though the teacher model's heat map does not capture the correct location, the student model can still identify it.
In this case, the student model might learned discriminative features from other samples in this area to localize the ground camera. 
This demonstrates the effectiveness of 
\newtext{adapting the student model to the target area by our knowledge distillation process.}

\subsection{Analysis of Prediction Errors after KD}
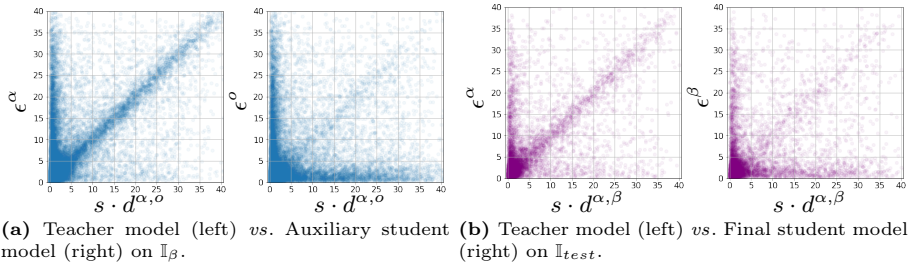
\begin{figure*}[ht]
    \centering
    \begin{subfigure}[t]{0.49\linewidth}
    \setlength{\abovecaptionskip}{0pt}
    \setlength{\belowcaptionskip}{0pt}
        \begin{minipage}[t]{0.49\textwidth}
    \begin{tikzpicture}
        \node (img) {\includegraphics[width=0.87\linewidth]{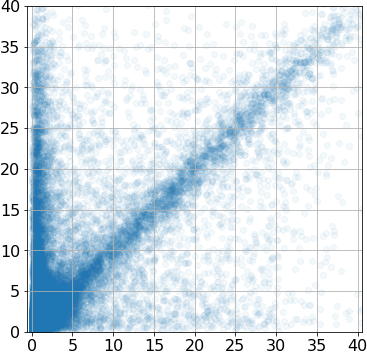}};
        \node[below=of img, node distance=0cm, yshift=1.2cm] {$ s\cdot d^{\source, o}$};
        \node[left=of img, node distance=0cm, rotate=90, anchor=center,yshift=-0.9cm] {$\epsilon^{\source}$};
    \end{tikzpicture}
    \end{minipage}%
    \begin{minipage}[t]{0.49\textwidth}
    \begin{tikzpicture}
        \node (img) {\includegraphics[width=0.87\linewidth]{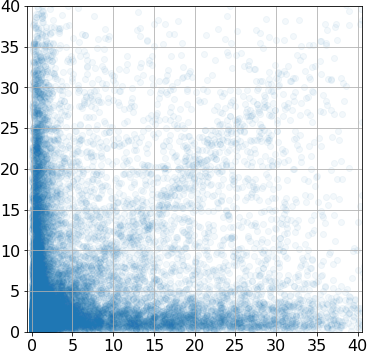}};
        \node[below=of img, node distance=0cm, yshift=1.2cm] {$ s \cdot d^{\source, o}$};
        \node[left=of img, node distance=0cm, rotate=90, anchor=center,yshift=-0.9cm] {$\epsilon^{o}$};
    \end{tikzpicture}
    \end{minipage}%
    \caption{Teacher model (left) \vs Auxiliary student model (right) on $\imageset_\target$. }
    \label{subfig:auxiliary_student}
    \end{subfigure}
    \begin{subfigure}[t]{0.49\linewidth}
    \setlength{\abovecaptionskip}{0pt}
    \setlength{\belowcaptionskip}{0pt}
    \begin{minipage}[t]{0.49\textwidth}
    \begin{tikzpicture}
        \node (img) {\includegraphics[width=0.87\linewidth]{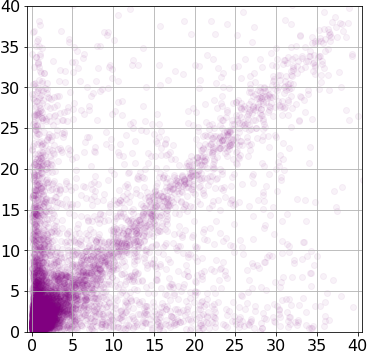}};
        \node[below=of img, node distance=0cm, yshift=1.2cm] {$ s\cdot d^{\source, \target}$};
        \node[left=of img, node distance=0cm, rotate=90, anchor=center,yshift=-0.9cm] {$\epsilon^{\source}$};
    \end{tikzpicture}
    \end{minipage}%
    \begin{minipage}[t]{0.49\textwidth}
    \begin{tikzpicture}
        \node (img) {\includegraphics[width=0.87\linewidth]{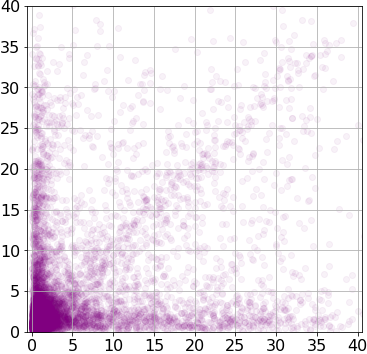}};
        \node[below=of img, node distance=0cm, yshift=1.2cm] {$ s \cdot d^{\source, \target}$};
        \node[left=of img, node distance=0cm, rotate=90, anchor=center,yshift=-0.9cm] {$\epsilon^{\target}$};
    \end{tikzpicture}
    \end{minipage}%
    \caption{Teacher model (left) \vs Final student model (right) on $\imageset_{test}$.}
    \label{subfig:final_student}
    \end{subfigure}
    \caption{CCVPE model, relation between error $\epsilon$ and change $d$ in predicted locations from teacher and student models on VIGOR. $\epsilon^{\source}$ / $\epsilon^{o}$ / $\epsilon^{\target}$: errors (m) of teacher model's / auxiliary student model's / final student model's predictions. $s\cdot d^{\source, o}$ / $s\cdot d^{\source, \target}$: the difference (m) between predicted locations of teacher and auxiliary / final student.
    }
    \label{fig:error_change_in_prediction}
\end{figure*}

Following the visual examples, we now analyze the overall statistical relation between the model prediction errors, and the change in predicted locations after knowledge distillation.
Figure~\ref{fig:error_change_in_prediction} plots this relation for CCVPE. 
The results for GGCVT are included in our Supplementary Material.

First, we confirm that potential outliers can indeed be identified by the amount of difference between the predicted locations of a teacher and its auxiliary student model in Figure~\ref{subfig:auxiliary_student} left.
We see there is a large portion of samples located around the diagonal line, \ie $\epsilon^\source = s\cdot d^{\source, o}$.
Most samples in $\imageset_\source$ with large change $d^{\source, o}$ in predicted location indeed obtained a large error $\epsilon^\source$ for the teacher model's prediction.
Next, Figure~\ref{subfig:auxiliary_student} right
shows how the difference in location correlates with the prediction error of the auxiliary student.
There are more samples being scattered at the bottom of the plot,
implying many wrong predictions of the teacher model have already been corrected.
Still, our ablation study will demonstrate that 
using the auxiliary student model directly as a new teacher for a final student model does not work as well as using it for outlier detection.
Note that the (less prominent) diagonal line now indicates errors introduced by the auxiliary student model. 
%
Lastly, 
we validate that the final student model reduces the localization error compared to the teacher model on the target test set $\imageset_{test}$ in Figure~\ref{subfig:final_student}.
Comparing the left plot to the right plot, we observe 
a similar trend as for the auxiliary student model before,
namely that the many samples with high teacher error in the left plot now obtain low student error in the right plot.

\subsection{Entropy Minimization}
We also tested entropy minimization~\cite{grandvalet2004semi} for the CCVPE model on the VIGOR dataset
as an alternative domain adaptation technique.
We tuned the strength of entropy minimization on predicted heat maps of training samples from the target area but found that stronger entropy minimization always leads to higher localization errors.
The best performance appears when no entropy minimization is applied.
Therefore, simply exposing the model to the images from the target area and enforcing the confidence of outputs is not sufficient for improving cross-view localization across areas.
We also observe that entropy minimization makes all heat maps sharper than direct generalization, but does not help the model resolve wrong modes.
Our proposed knowledge self-distillation instead reduces uncertainty by filtering out unreliable samples.

\subsection{Ablation Study}
\begin{figure*}[t]
    \centering
    \begin{minipage}{0.66\linewidth}
        \includegraphics[width=0.485\linewidth]{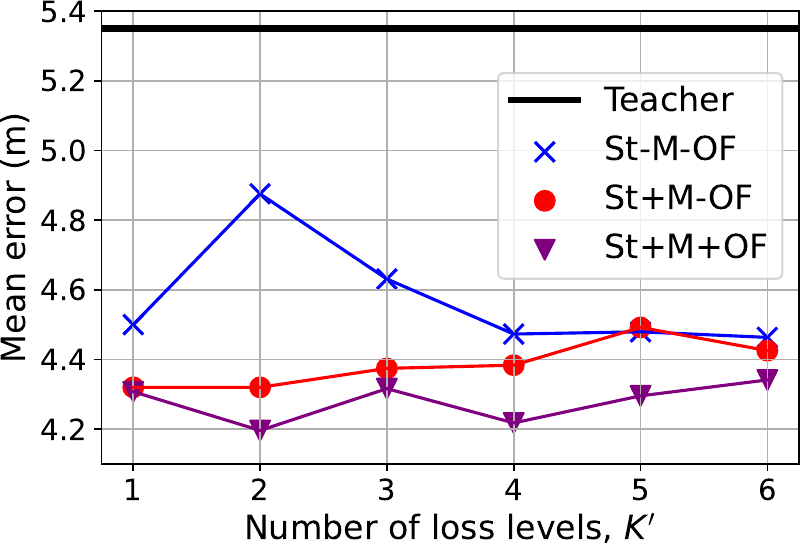}
    \includegraphics[width=0.485\linewidth]{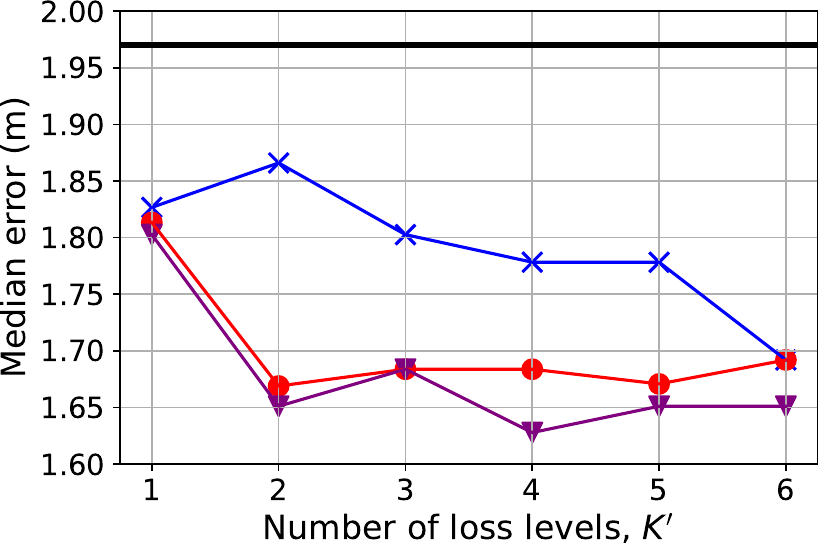}
    \captionof{figure}{ Ablation study on the proposed mode-based pseudo ground truth, outlier filtering, and different levels for coarse-only supervision in our teacher-student KD using CCVPE. }
    \label{fig:ablation_CCVPE_GGCVT_VIGOR}
    \end{minipage}
    \hfill
    \begin{minipage}{0.32\linewidth}
        \includegraphics[width=\linewidth]{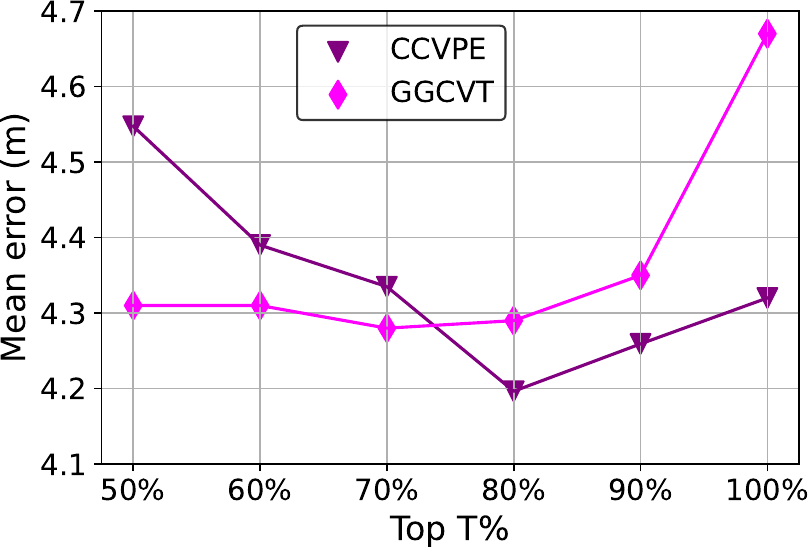}
    \captionof{figure}{ Effect of $T$ in the proposed outlier filtering. $100\%$ means no outlier filtering.
    }
    \label{fig:ablation_CCVPE_GGCVT_VIGOR_T}
    \end{minipage}
\end{figure*}

An extensive ablation study is conducted to validate the effectiveness of our proposed designs.
We denote the following:
\definemodel{Teacher} (baseline): directly deploy the teacher model $\mathcal{M}^\source$ in the target area.
\definemodel{St-M-OF}: student model trained using teacher's heat maps, no mode-based pseudo ground truth, no outlier filtering.
\definemodel{St+M-OF}: student model trained using mode-based pseudo ground truth, no outlier filtering.
\definemodel{St+M+OF} (proposed): student model trained using mode-based pseudo ground truth with outlier filtering, \ie \newtext{the model} $\mathcal{M}^{\target}$.



The performance of these ablation variants when supervising different levels of student predictions of the CCVPE is shown in Figure~\ref{fig:ablation_CCVPE_GGCVT_VIGOR}. 
It can be seen that the proposed mode-based pseudo ground truth (+M) and outlier filtering (+OF) both improve the performance and the final version, St+M+OF, achieves the best results, no matter how many prediction levels of the student model are supervised. 
For CCVPE student models, supervising the first $K'=2$ and $K'=4$ levels have similar localization performance overall.
Since $K'=2$ gives the lowest mean error, we use it in our final setting.
We also tuned $K'$ for GGCVT and found that supervising all three levels, \ie $K'=3$ gives the best results. 
The effectiveness of the proposed mode-based pseudo ground truth (+M) and outlier filtering (+OF) on GGCVT is verified in Table~\ref{tab:GGCVT_ablation}.
\newtext{When not using any of the proposed designs,
\ie GGCVT student model follows Best Teacher Distillation~\cite{zhang2021self}, the student's performance (5.34~m) is worse than the Teacher's (5.16~m).
This highlights the importance of reducing uncertainty and removing outliers in teacher's predictions.}
Additionally, we also tried directly using the predictions of the auxiliary student as pseudo ground truth to train the final student model (similar to iterative knowledge self-distillation~\cite{furlanello2018born}), denoted as St+M+A in Table~\ref{tab:GGCVT_ablation}.
However, it does not perform better than using the auxiliary student model for outlier filtering.

\begin{table}[!t]
    \caption{Ablation study for GGCVT. \textbf{Best in bold.}}
    \setlength{\tabcolsep}{3pt}
    \centering
    \begin{tabular}{cccccc}
    \toprule
    Error (m) & Teacher & St-M-OF & St+M-OF & St+M+A & St+M+OF \\
    \hline
    Mean & 5.16 & 5.34 & 4.67 & 4.54 & \textbf{4.28}\\
    \hline
    Median & 1.40 & 1.48 & 1.32 & 1.55 & \textbf{1.28}\\
    \bottomrule
    \end{tabular}
    \label{tab:GGCVT_ablation}
\end{table}

Figure~\ref{fig:ablation_CCVPE_GGCVT_VIGOR_T} shows the ablation study results on different percentage values $T$ in our outlier detection.
The best CCVPE and GGCVT student models appear at $T=80\%$ and $T=70\%$.
In general, there is a trade-off between the quality and quantity of data.
When too little data is kept, there is a risk of model overfitting. 
Filtering out some detected outliers ($20\%\sim 30\%$) improves the quality of the data and can result in better 
model performance.
This suggests that, in practice, blindly increasing the data amount without guaranteeing its quality might negatively influence models' performance.



\section{Conclusion}

This paper focuses on improving the localization performance of a pre-trained fine-grained cross-view localization model in a new target area without \newtext{any fine} ground truth. 
We have proposed a knowledge self-distillation-based weakly-supervised learning approach that only requires a set of ground-aerial image pairs from the target area.
Extensive experiments were conducted to study how to generate appropriate pseudo ground truth for student model training.
We found that selecting the predominant mode in the teacher model's predictions is better than directly using the output heat maps.
Furthermore, supervising coarse-level predictions of a student model using the down-sampled teacher model's high-resolution predictions can suppress the positional noise and might lead to a slight boost in the student model's performance.
Last but not least, we demonstrate that unreliable target domain samples can be filtered out by comparing predicted locations from teacher and student models, which motivates using an auxiliary student model to curate the data. 
Training a final student model on the filtered data further improves the localization accuracy.
Our proposed approach has been validated on two state-of-the-art methods on two benchmarks. It achieves a consistent and considerable performance boost over the previous standard that directly deploys the trained model in the new target area.

\title{Supplementary Material} 
\author{}
\institute{}
\maketitle

In this supplementary material, we provide the following information to support the main paper:
\begin{enumerate}[label=\Alph*]
    \item Supervised Finetuning with Noisy Ground Truth.
    \item Domain Adaptation by Entropy Minimization.
    \reviewersugg{\item Domain Adaptation by Other Pseudo Label-based Approaches.}
    \item Analysis of Prediction Errors after KD for GGCVT~\cite{shi2023boosting}.
    \item Error Distribution of Teacher and Final Student Model.
    \item Extra Qualitative Results.
    \reviewersugg{\item T-SNE Feature.}
    \item Potential Negative Impact.
    \item Limitations.
\end{enumerate}

\section*{A. Supervised Finetuning with Noisy Ground Truth}
As mentioned in Section 4.5 of the main paper, when the ground truth for cross-area supervised finetuning contains errors, the finetuned model has large test errors.

\begin{figure}
    \centering
    \includegraphics[width=0.4\linewidth]{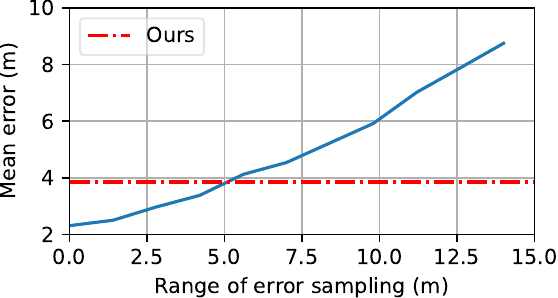}
    \hspace{0.5cm}
    \includegraphics[width=0.4\linewidth]
    {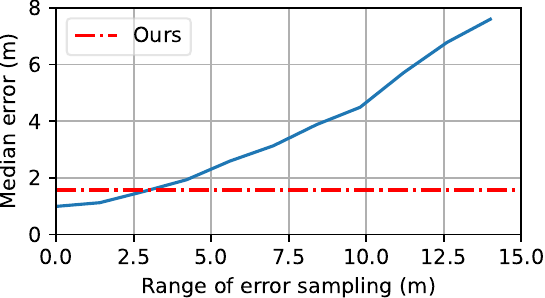}
    \caption{VIGOR test set errors (vertical axis) of CCVPE models fintuned on noisy ground truth. The horizontal axis denotes the upper bound for error sampling.}
    \label{fig:train_on_noisy_GT}
\end{figure}

In our experiments, offsets were sampled randomly and uniformly (in both the north-south and east-west directions) within a defined range for each ground-level image in the cross-area training set prior to fine-tuning.
These offsets were then applied to shift the ground truth locations of the training images. 
As Figure~\ref{fig:train_on_noisy_GT} demonstrates, inaccuracies in the ground truth markedly affect the localization precision of the fine-tuned model. 
For the supervised fine-tuned model to outperform the model trained with our weakly-supervised learning approach in terms of both mean and median test errors, the maximum permissible error in the ground truth for each direction should be under approximately $2.5$ m.
In practice, acquiring ground truth with this level of accuracy on a large scale is difficult, as standard GNSS positioning does not meet this requirement~\cite{benmoshe2011urbangnss}. 
Instead, our proposed method requires only ground-aerial image pairs, making it a more scalable solution in practice.

\section*{B. Domain Adaptation by Entropy Minimization}
As noted in our main paper Section~4.7, we explore entropy minimization~\cite{grandvalet2004semi} as an alternative approach to adapt a model from the source domain to the target domain.
Entropy minimization is often used for semi-supervised domain adaptation~\cite{vu2019advent}.
In this setting, the model is trained with a combination of samples with ground truth labels from the source domain and unlabeled samples from the target domain.
When a source domain sample is presented, the model is trained using its default supervised learning loss $\mathcal{L_M}$.
When the input is from the target domain, the training objective is to minimize the entropy of the output prediction using an entropy minimization loss $\mathcal{L}_{EM}$.

\begin{figure}[ht]
    \centering
    \begin{tikzpicture}
        \node (img) {\includegraphics[width=0.4\linewidth]{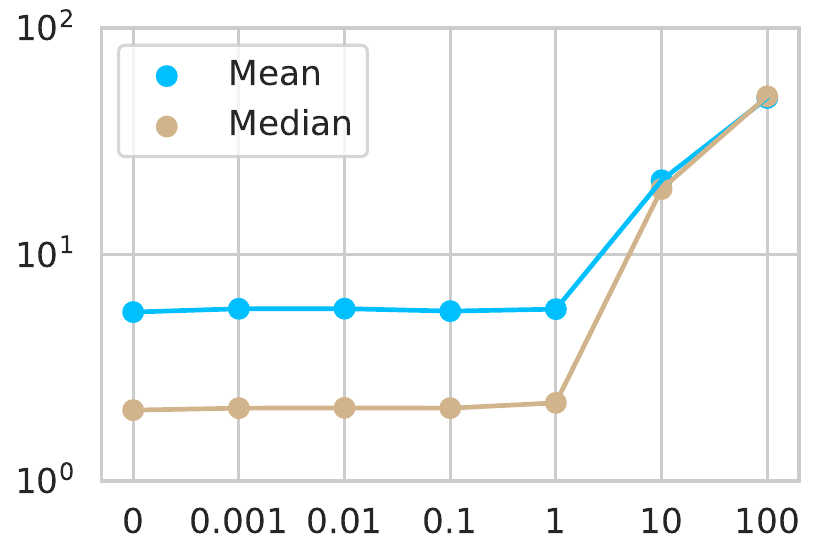}};
        \node[below=of img, node distance=0cm, xshift=0.25cm, yshift=1.2cm] {\scriptsize Entropy minimization weight, $\omega$};
        \node[left=of img, node distance=0cm, rotate=90, anchor=center,yshift=-1cm] {\scriptsize Error, $\epsilon$ (m)};
    \end{tikzpicture}
    \caption{Errors of CCVPE models with different entropy minimization weights $\omega$ on VIGOR validation set.}
    \label{fig:entropy_minization}
\end{figure}

We train a CCVPE model~\cite{xia2023convolutional} on a combination of VIGOR source and target domain data using loss $\mathcal{L}_{final}$,
\begin{align}
    \mathcal{L}_{final} = \begin{cases}
    \mathcal{L_M}(\mathcal{M}(G, A), \hat{y}),& \text{if } \{G,A\}\in\imageset_\source, \hat{y} \in \labelset_\source,\\
    \omega \cdot \mathcal{L}_{EM}(H_K),              & \text{if } \{G,A\}\in\imageset_\target.
    \end{cases}
    \label{eq:EM_final_loss}
\end{align}
In Equation~\ref{eq:EM_final_loss}, $\mathcal{L_M}$ is the default supervised learning loss of CCVPE~\cite{xia2023convolutional}, $H_K$ is the final output heat map of the model $\mathcal{M}$ on image pair $\{G,A\}$, and $\omega$ is a hyperparameter that weighs the entropy minimization loss $\mathcal{L}_{EM}$. As in~\cite{vu2019advent}, we calculate the pixel-wise Shannon Entropy~\cite{shannon2001mathematical} in the dense output, and then use the sum of all pixel-wise entropy as our $\mathcal{L}_{EM}$,
\begin{align}
    \mathcal{L}_{EM}(H_K) = -\sum_{u,v} H_K(u,v) \cdot \log(H_K(u,v)),
\end{align}
$H_K(u,v)$ denotes the value at each location in the output heat map $H_K$.

\textbf{We tuned $\omega$ and found that joint training with entropy minimization always hurts the model performance.}
As shown in Figure~\ref{fig:entropy_minization}, the mean and median error on the validation set (target area) increases when the model is trained using a larger weight $\omega$, and the best model appears when $\omega=0$, equivalent to direct generalization of a model trained in a supervised manner on only source domain images.

For completeness, we also tried directly finetuning a pre-trained model from the source domain on images from the target domain using entropy minimization (no joint supervised training with source domain samples).
Since the model failed completely, we did not include the plots.

\begin{figure*}[t]
    \centering
    \begin{minipage}{\linewidth}
    \centering
    \includegraphics[align=c, width=0.4\linewidth]{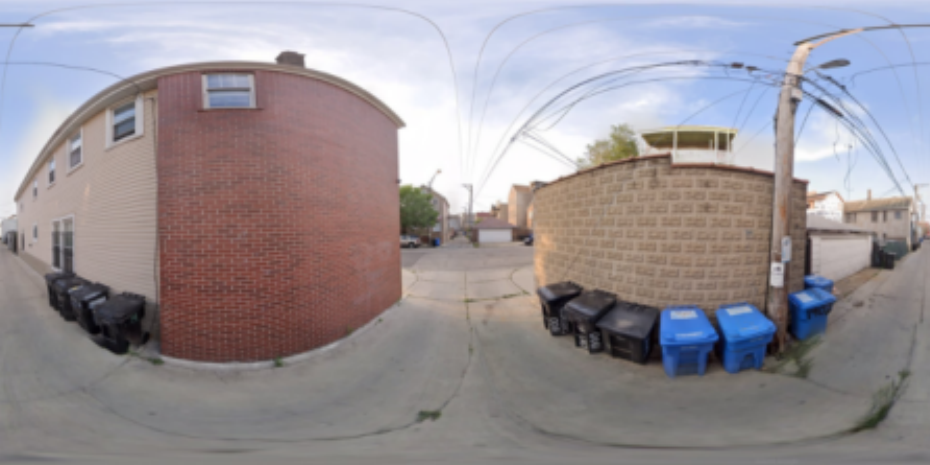} \\ 
    \includegraphics[align=c, width=0.24\linewidth]{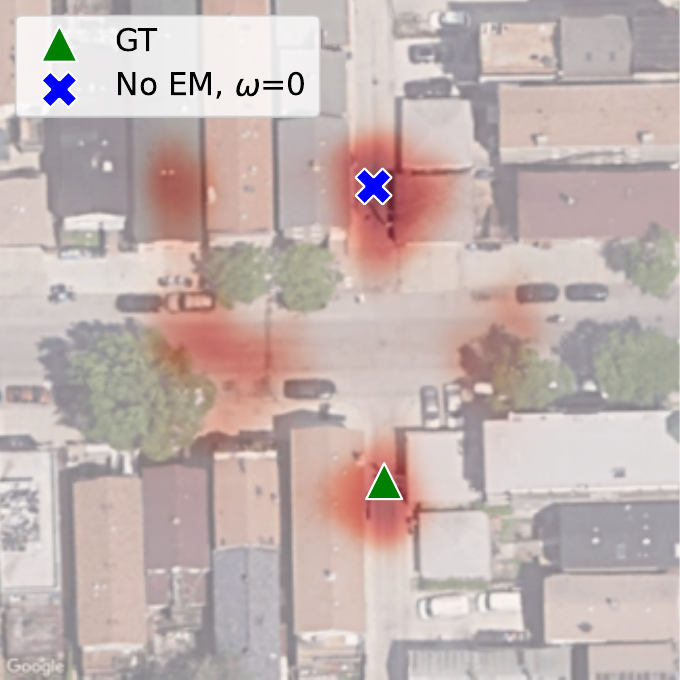}
    \includegraphics[align=c, width=0.24\linewidth]{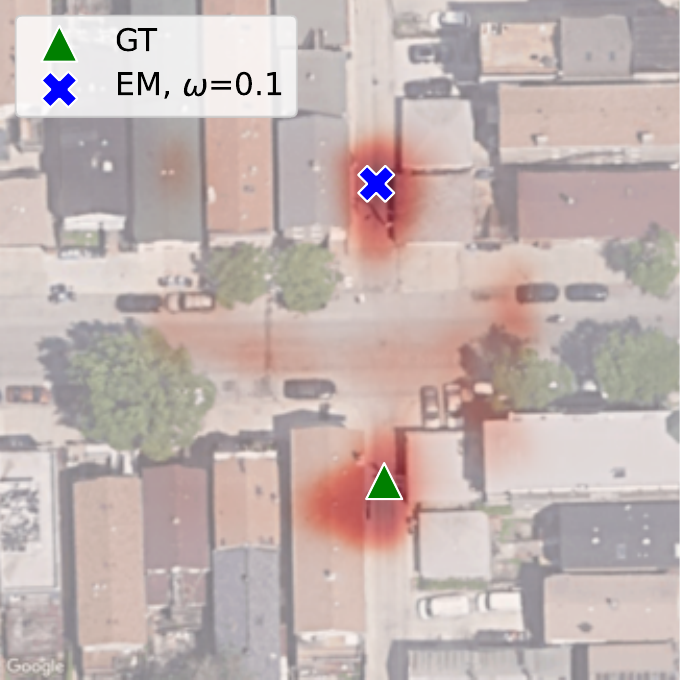}
    \includegraphics[align=c, width=0.24\linewidth]{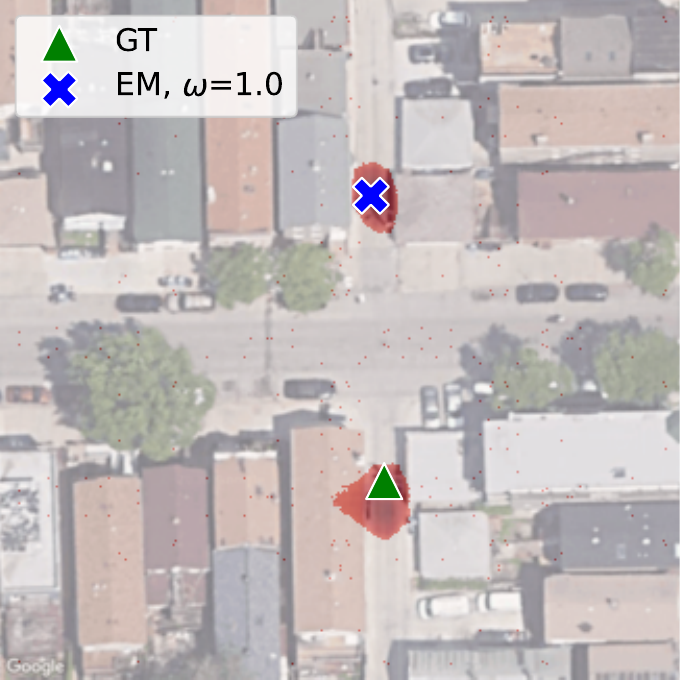}
    \includegraphics[align=c, width=0.24\linewidth]{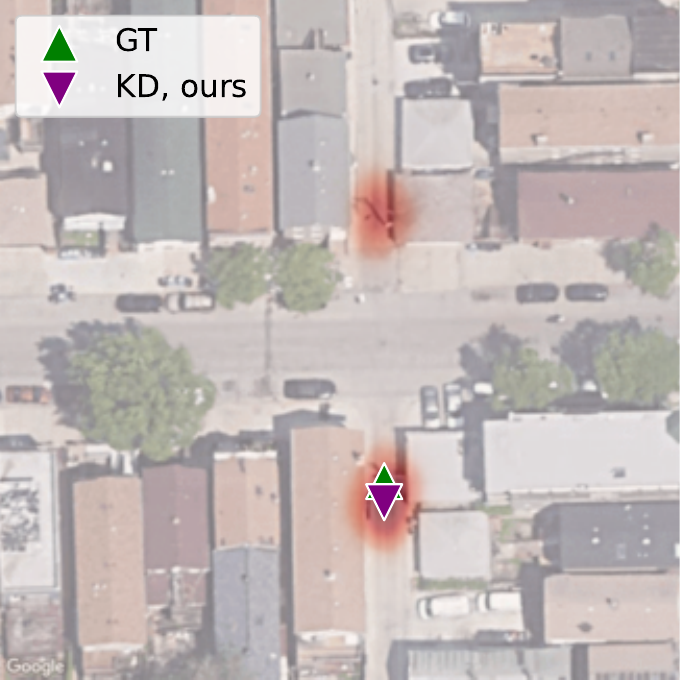}
    \\
    \includegraphics[align=c, width=0.4\linewidth]{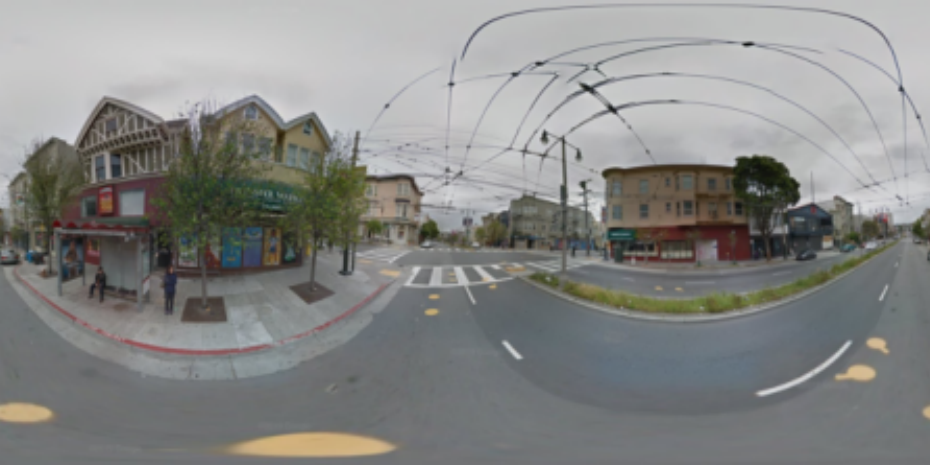} \\
    \includegraphics[align=c, width=0.24\linewidth]{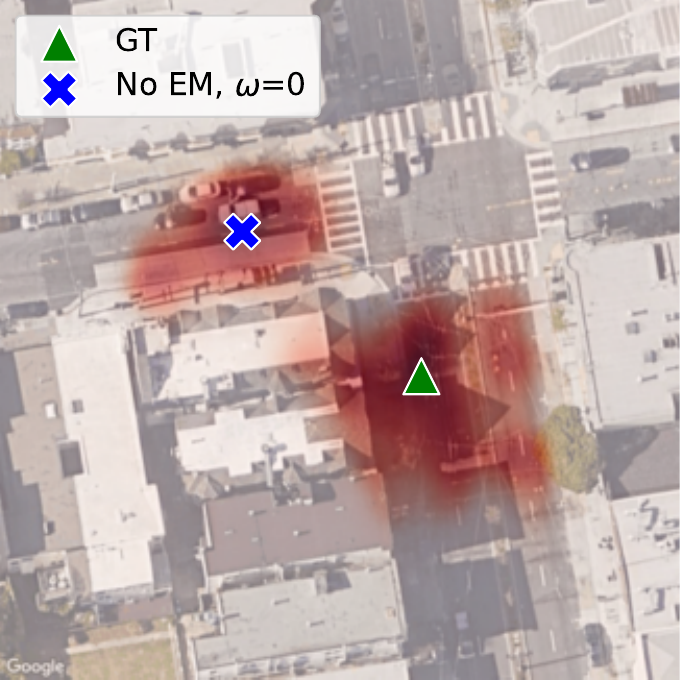}    
    \includegraphics[align=c, width=0.24\linewidth]{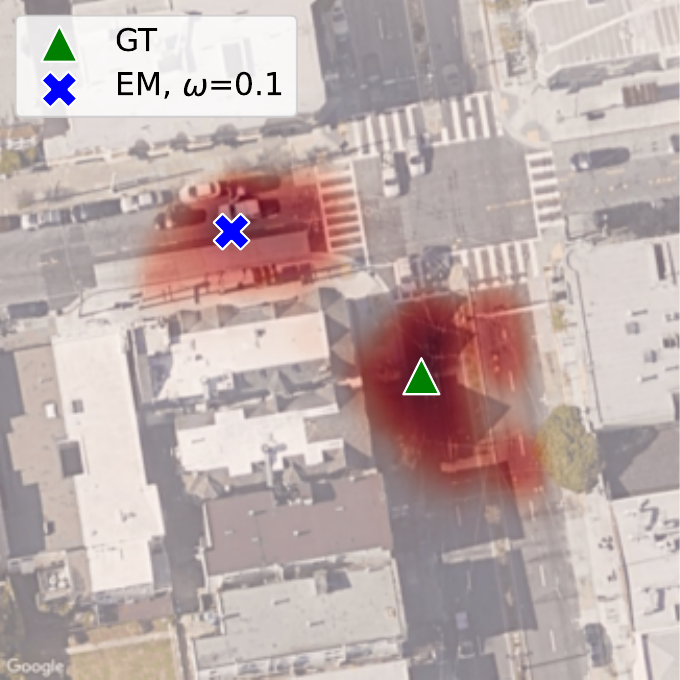}    
    \includegraphics[align=c, width=0.24\linewidth]{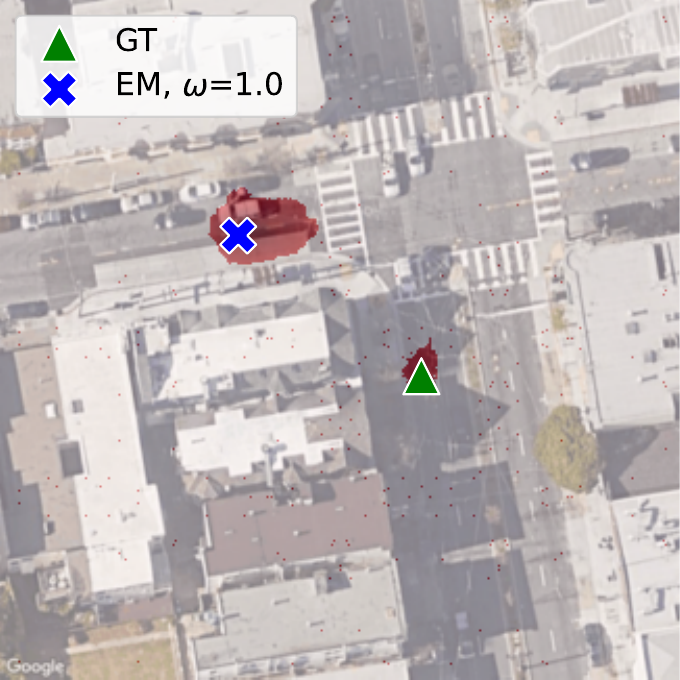}
    \includegraphics[align=c, width=0.24\linewidth]{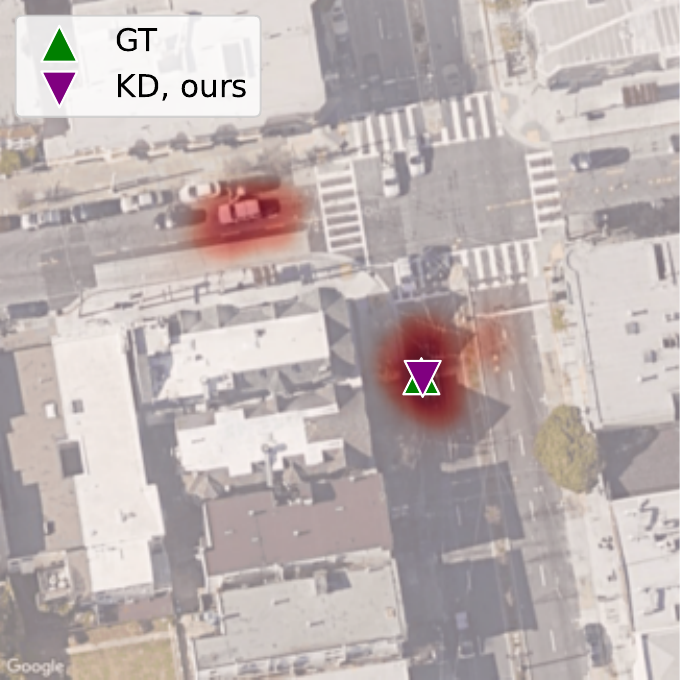}
    \end{minipage}
    \caption{Adapting a CCVPE model to the target domain with different methods. Results on the VIGOR test set. Comparison between direct generalization (No EM, $\omega=0$), different entropy minimization weights (EM, $\omega=0.1$ and EM, $\omega=1.0$), and our proposed knowledge self-distillation (KD, ours).
    The red color denotes the localization probability (a darker color means a higher probability).}
    \label{fig:KD_vs_EM}
\end{figure*}

Entropy minimization simply encourages the heat map to be sharper in the target area.
Therefore, it does not resolve multi-modal uncertainty.
As shown in Figure~\ref{fig:KD_vs_EM}, compared to direct generalization, training with entropy minimization makes the red region in the heat map smaller, but the peak of the heat map stays in the same mode in the multi-modal distribution.
Instead, our proposed knowledge self-distillation adapts the model to the target domain by explicitly encouraging the model to disambiguate multiple modes using the proposed single-modal pseudo ground truth.
As a result, our proposed method can correct the wrong mode and also reduce uncertainty.

\section*{C. Domain Adaptation by Other Pseudo Label-based Approaches}
\reviewersugg{
Our proposed Coarse-only Supervision uses the model's highest resolution output to supervise low-resolution ones.
Alternatively, we also studied fusing the outputs at different levels to generate supervision signals.
}

\reviewersugg{
Similar to~\cite{ji2021refine}, we fuse information in both top-down and bottom-up directions to generate pseudo ground truth at each level for the student model.
We achieved this by up/downsampling teacher's matching volumes at different levels and fusing them with averaging.
The error of the resulting student (4.49 m) is larger than ours (3.85 m) and the teacher model (4.38 m).
We hypothesize that for localization, fine-grained high-resolution heatmaps can help supervise low-resolution maps, but not vice versa,
which may be why \cite{ji2021refine}'s top-down + bottom-up approach does not work for our task.
}

\reviewersugg{
As an alternative to our proposed outlier filtering, we also tried an uncertainty-based outlier filtering approach while keeping other proposed modules unchanged.
Similar to~\cite{zhou2022uncertainty,wang2021uncertainty,litrico2023guiding}, we use the entropy of teacher's output heat maps as a measure of their uncertainty.
The teacher's heatmaps are ranked based on their entropy and we use the most certain $T\%$ for student training.
For a fair comparison, CCVPE uses top $80\%$ and GGCVT uses top $70\%$ (same as in our outlier detection). 
The resulting models have higher errors (CCVPE/GGCVT: 4.17/4.52 m) than ours (3.85/4.34 m).
Entropy-based methods do not consider the spatial order of classes, e.g. a two-mode heatmap with 1 m between two modes will have the same entropy as a two-mode heatmap with 10 m between modes.
However, the latter results in larger errors.
}

\section*{D. Analysis of Prediction Errors after KD for GGCVT}
\begin{figure}[ht]
    \centering
    \setlength{\abovecaptionskip}{0pt}
    \setlength{\belowcaptionskip}{0pt}
    \begin{subfigure}[t]{0.48\linewidth}
    \setlength{\abovecaptionskip}{0pt}
    \setlength{\belowcaptionskip}{0pt}
        \begin{minipage}[t]{0.48\textwidth}
    \begin{tikzpicture}
        \node (img) {\includegraphics[width=0.87\linewidth]{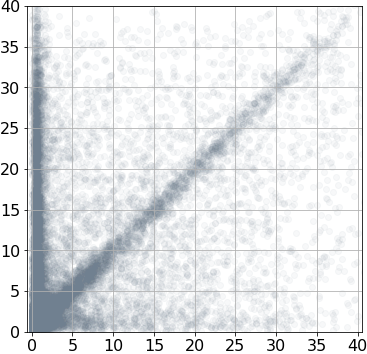}};
        \node[below=of img, node distance=0cm, yshift=1.2cm] {$ s\cdot d^{\source, o}$};
        \node[left=of img, node distance=0cm, rotate=90, anchor=center,yshift=-0.9cm] {$\epsilon^{\source}$};
    \end{tikzpicture}
    \end{minipage}%
    \begin{minipage}[t]{0.48\textwidth}
    \begin{tikzpicture}
        \node (img) {\includegraphics[width=0.87\linewidth]{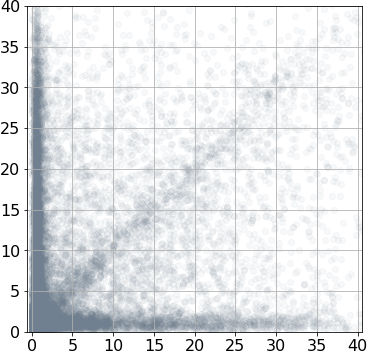}};
        \node[below=of img, node distance=0cm, yshift=1.2cm] {$ s \cdot d^{\source, o}$};
        \node[left=of img, node distance=0cm, rotate=90, anchor=center,yshift=-0.9cm] {$\epsilon^{o}$};
    \end{tikzpicture}
    \end{minipage}%
    \caption{Teacher (left) \vs Auxiliary student (right) models on $\imageset_\target$ }
    \label{subfig:auxiliary_student_GGCVT}
    \end{subfigure}
    \begin{subfigure}[t]{0.48\linewidth}
    \setlength{\abovecaptionskip}{0pt}
    \setlength{\belowcaptionskip}{0pt}
    \begin{minipage}[t]{0.48\textwidth}
    \begin{tikzpicture}
        \node (img) {\includegraphics[width=0.87\linewidth]{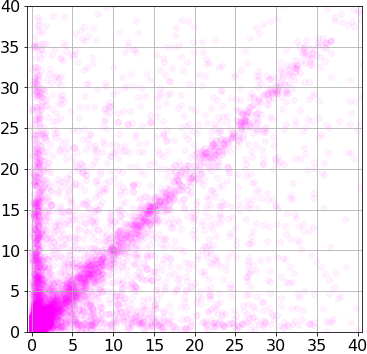}};
        \node[below=of img, node distance=0cm, yshift=1.2cm] {$ s\cdot d^{\source, \target}$};
        \node[left=of img, node distance=0cm, rotate=90, anchor=center,yshift=-0.9cm] {$\epsilon^{\source}$};
    \end{tikzpicture}
    \end{minipage}%
    \begin{minipage}[t]{0.48\textwidth}
    \begin{tikzpicture}
        \node (img) {\includegraphics[width=0.87\linewidth]{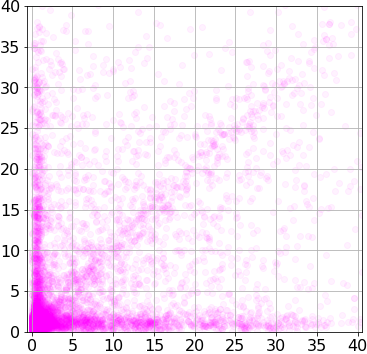}};
        \node[below=of img, node distance=0cm, yshift=1.2cm] {$ s \cdot d^{\source, \target}$};
        \node[left=of img, node distance=0cm, rotate=90, anchor=center,yshift=-0.9cm] {$\epsilon^{\target}$};
    \end{tikzpicture}
    \end{minipage}%
    \caption{Teacher (left) \vs Final student (right) models on $\imageset_{test}$}
    \label{subfig:final_student_GGCVT}
    \end{subfigure}
    \caption{GGCVT model, relation between error $\epsilon$ and change $d$ in predicted locations from teacher and student models on VIGOR. $\epsilon^{\source}$ / $\epsilon^{o}$ / $\epsilon^{\target}$: errors (m) of teacher model's / auxiliary student model's / final student model's predictions. $s\cdot d^{\source, o}$ / $s\cdot d^{\source, \target}$: the difference (m) between predicted locations of teacher and auxiliary / final student.
    }
    \label{fig:error_change_in_prediction_GGCVT}
\end{figure}

Similar to the analysis of the predictions of CCVPE in our main paper Section 4.6, we here provide the overall statistical relation between the GGCVT's prediction errors and the change in its predicted locations after knowledge distillation.
\textbf{Overall, we observe the same trend for GGCVT as we had for CCVPE in the main paper, see Figure~\ref{fig:error_change_in_prediction_GGCVT}.}

First, a strong correlation between the teacher model's prediction errors and the amount of difference between the predicted locations of a teacher and its auxiliary student model is observed from the diagonal line in Figure~\ref{subfig:auxiliary_student_GGCVT} left.
This again confirms that the outliers in the teacher's prediction can be identified by measuring the changes in the predicted location after knowledge self-distillation, no matter what the localization backbone is, demonstrating the effectiveness of our proposed outlier filtering.

Figure~\ref{subfig:auxiliary_student_GGCVT} right plot has many scattered points along the horizontal axis, representing the predictions that are corrected by the auxiliary student model.
The diagonal line in this plot then shows the samples in which the auxiliary student model introduced an error in its predictions, i.e. the correct teacher's predictions being moved to a wrong location or the wrong teacher's predictions being moved to another wrong location.

On the VIGOR test set $\imageset_{test}$, Figure~\ref{subfig:final_student_GGCVT} validated that the final GGCVT student model reduces the error of its teacher, as shown by the less prominent diagonal line and more points along the horizontal axis in the right plot compared to those in the left plot.

\section*{E. Error Distribution of Teacher and Final Student Model}

\begin{figure}[ht]
    \centering
    \begin{subfigure}{0.40\textwidth}
    \begin{tikzpicture}
        \node (img) {\includegraphics[width=0.94\linewidth]{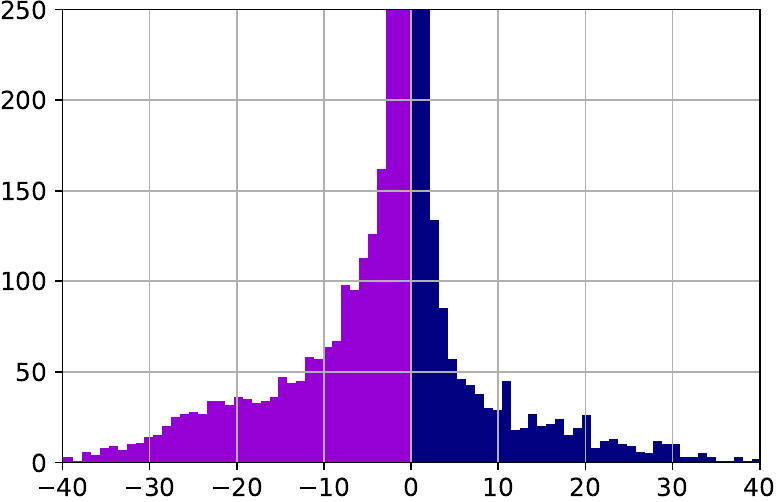}};
        \node[below=of img, node distance=0cm, yshift=1.2cm] {\scriptsize $\Delta \epsilon = \epsilon^{\target} - \epsilon^{\source}$ (m)};
        \node[left=of img, node distance=0cm, rotate=90, anchor=center,yshift=-0.9cm] {\scriptsize Number of samples};
        \draw (-1,0.7)
        node[text width=2cm,align=center,black,fill=none]{\scriptsize $\Delta \epsilon<0$ \\ Student is better};
        \draw (1.4,0.7)
        node[text width=2cm,align=center,black,fill=none]{\scriptsize $\Delta \epsilon>0$ \\ Teacher is better};
    \end{tikzpicture}
    \caption{CCVPE model}
    \end{subfigure}
    \hspace{3em}
    \begin{subfigure}{0.40\textwidth}
    \begin{tikzpicture}
        \node (img) {\includegraphics[width=0.94\linewidth]{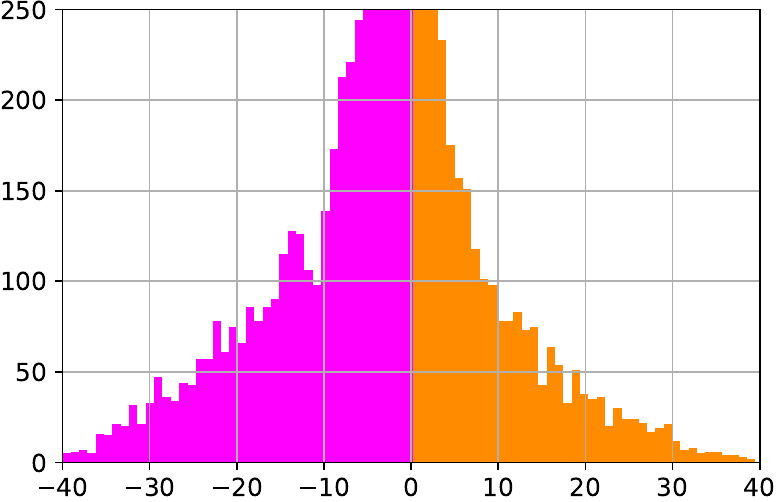}};
        \node[below=of img, node distance=0cm, yshift=1.2cm] {\scriptsize $\Delta \epsilon = \epsilon^{\target} - \epsilon^{\source}$ (m)};
        \node[left=of img, node distance=0cm, rotate=90, anchor=center,yshift=-0.9cm] {\scriptsize Number of samples};
        \draw (-1,0.7)
        node[text width=2cm,align=center,black,fill=none]{\scriptsize $\Delta \epsilon<0$ \\ Student is better};
        \draw (1.4,0.7)
        node[text width=2cm,align=center,black,fill=none]{\scriptsize $\Delta \epsilon>0$ \\ Teacher is better};
    \end{tikzpicture}
    \caption{GGCVT model}
    \end{subfigure}%
    \caption{Change in error between predictions of the teacher $\mathcal{M}^\source$ and those of the student model $\mathcal{M}^\target$ on VIGOR test set $\imageset_{test}$. Purple and Magenta region: The student model has smaller errors. Navy and Orange region: The teacher has smaller errors.
    }
    \label{fig:error_histogram}
\end{figure}

Next, we compare the error in predictions of the teacher model and that of the student model for both CCVPE and GGCVT on the VIGOR test set $\imageset_{test}$.
We calculate the error change after weakly-supervised knowledge self-distillation and visualize the statistics in Figure~\ref{fig:error_histogram}.
The left part of the two histograms (in purple and magenta) shows the samples that have a smaller error in the student model's prediction.
Similarly, the right part of the two histograms (in navy and orange) denotes the samples that the teacher model has a more accurate prediction.
Overall, we see that, for both CCVPE and GGCVT, there are more samples located in the left part.
\textbf{It demonstrates that the student model reduces the error for the majority of samples.}

\section*{F. Extra qualitative results of teacher and student models}
\begin{figure*}[ht]
    \centering
    \begin{minipage}{\linewidth}
    \centering
    \includegraphics[align=c, width=0.24\linewidth]{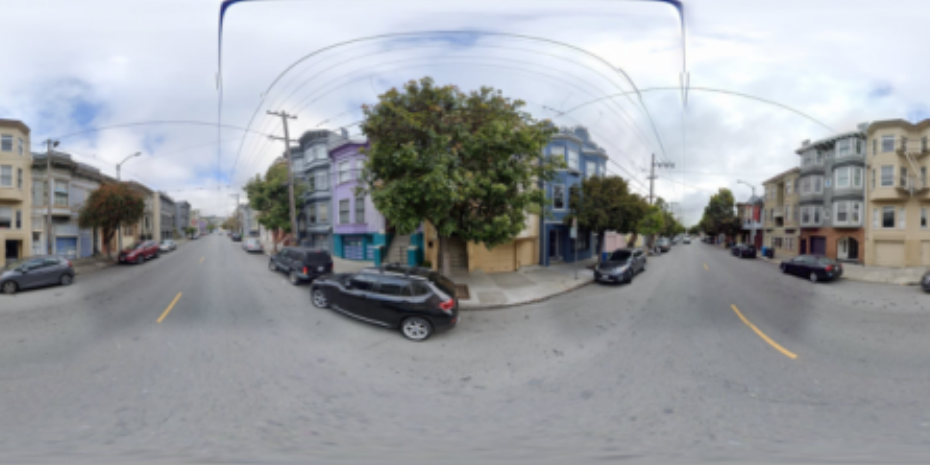}
    \includegraphics[align=c, width=0.24\linewidth]{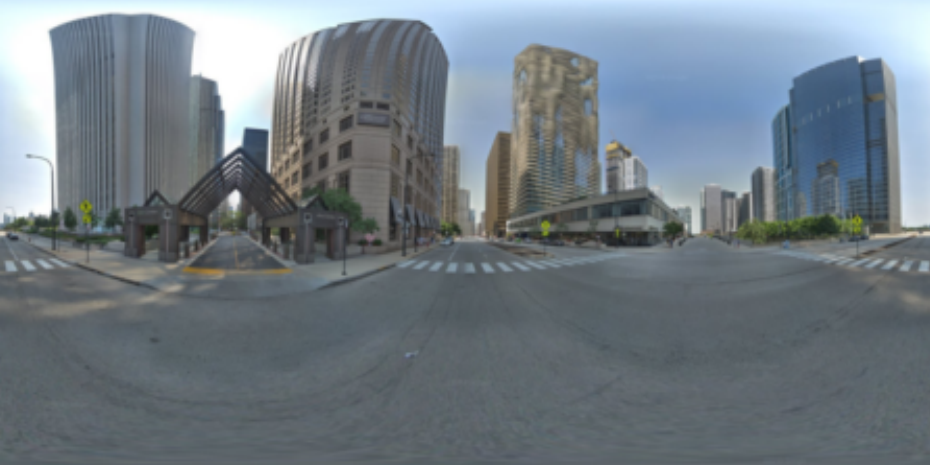}
    \includegraphics[align=c, width=0.24\linewidth]{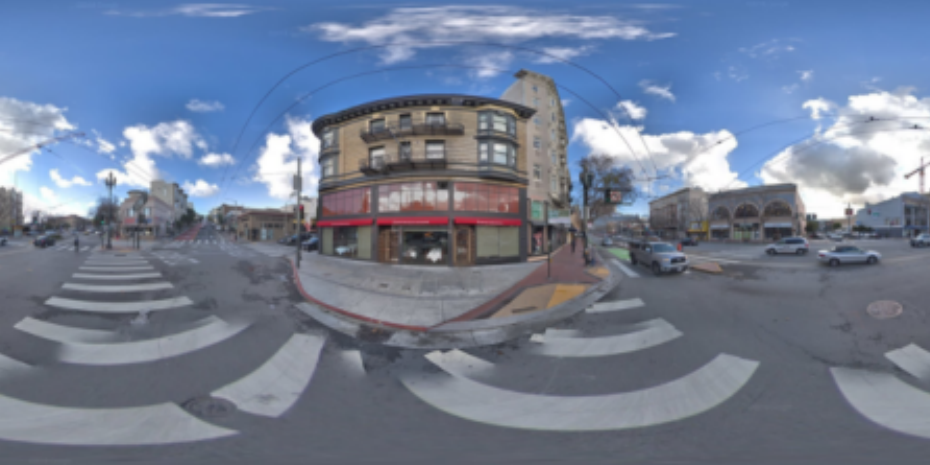}
    \includegraphics[align=c, width=0.24\linewidth]{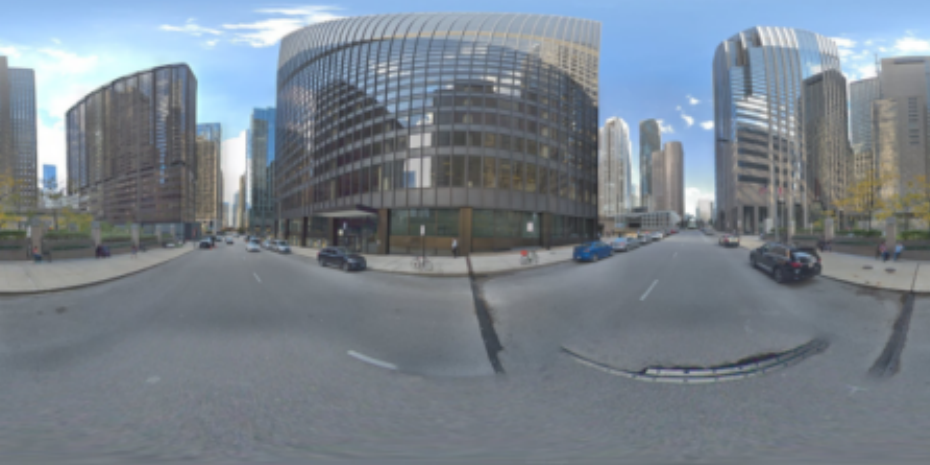} 
    \\
    \includegraphics[align=c, width=0.24\linewidth]{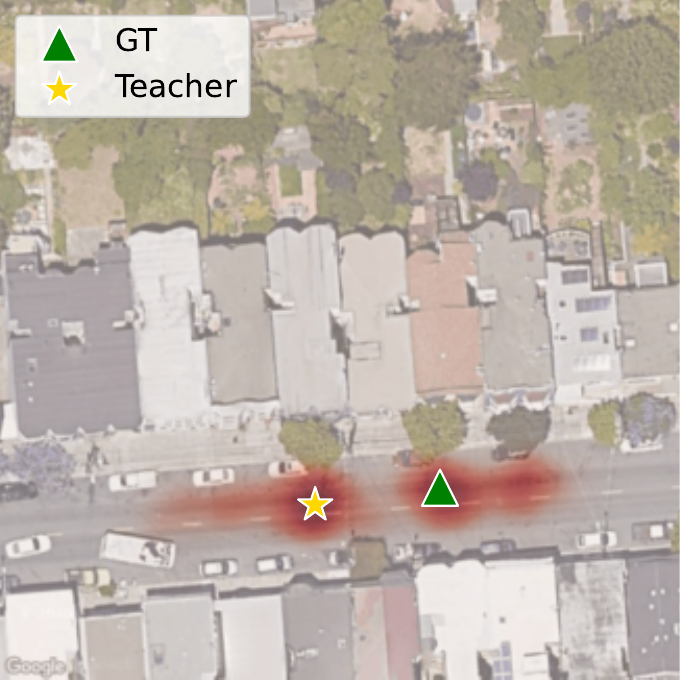}
    \includegraphics[align=c, width=0.24\linewidth]{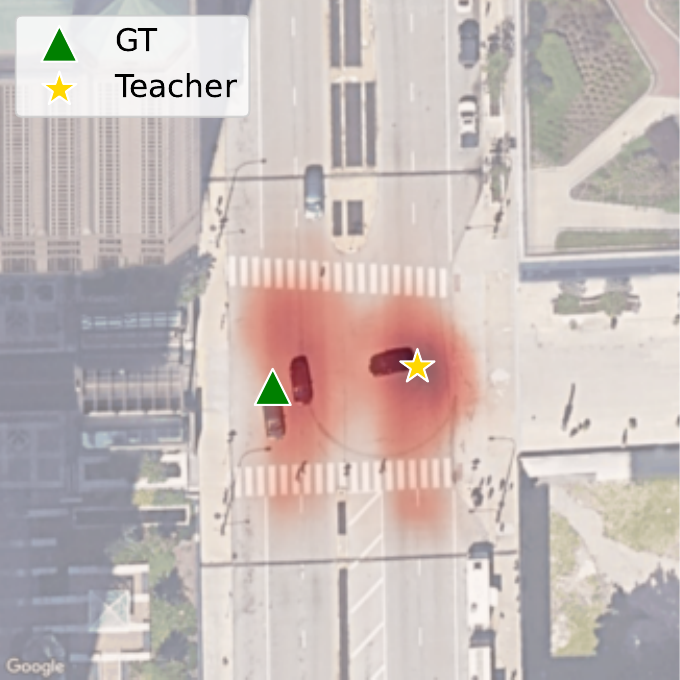}
    \includegraphics[align=c, width=0.24\linewidth]{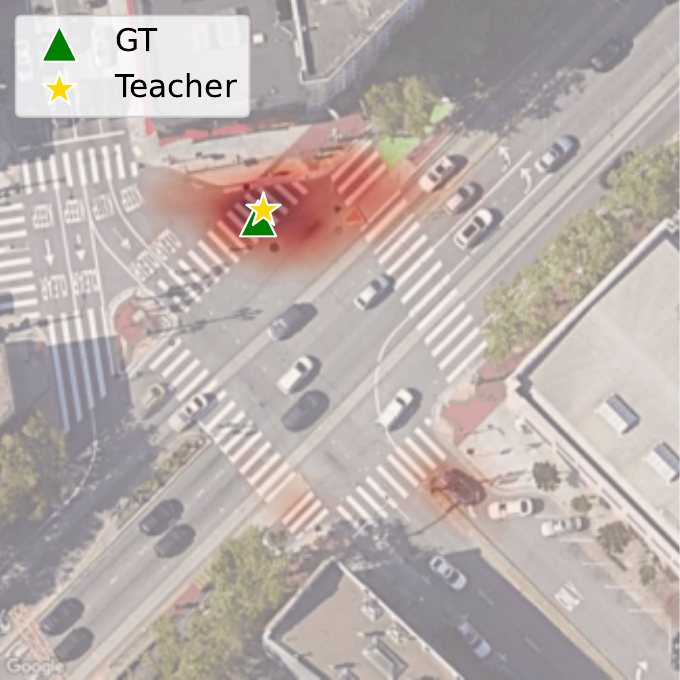}
    \includegraphics[align=c, width=0.24\linewidth]{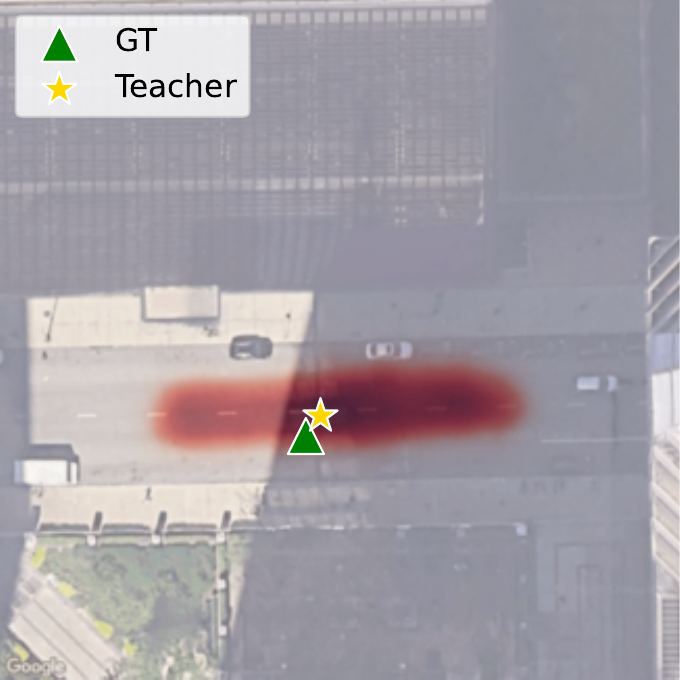}
    \\
    \includegraphics[align=c, width=0.24\linewidth]{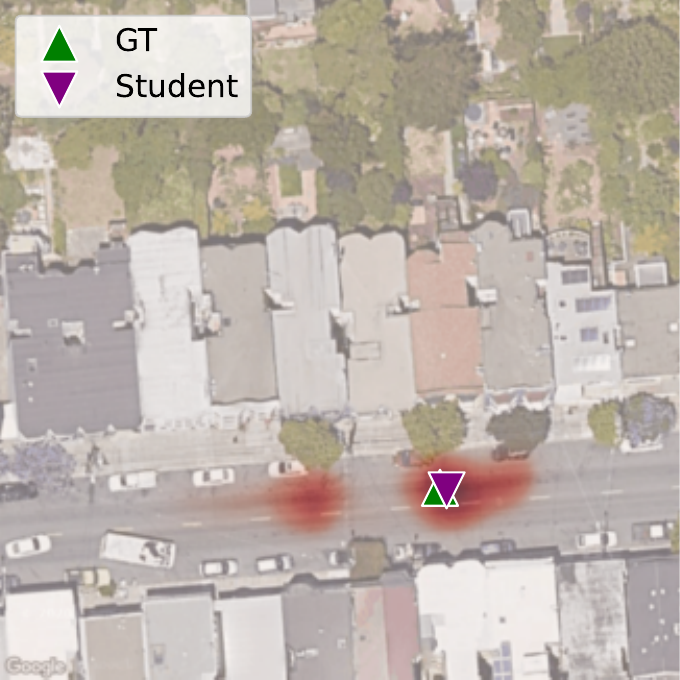}
    \includegraphics[align=c, width=0.24\linewidth]{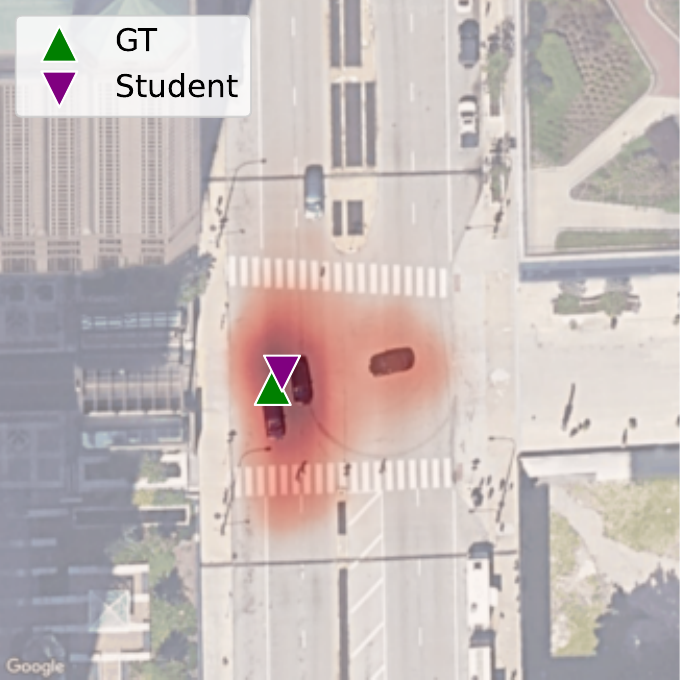}
    \includegraphics[align=c, width=0.24\linewidth]{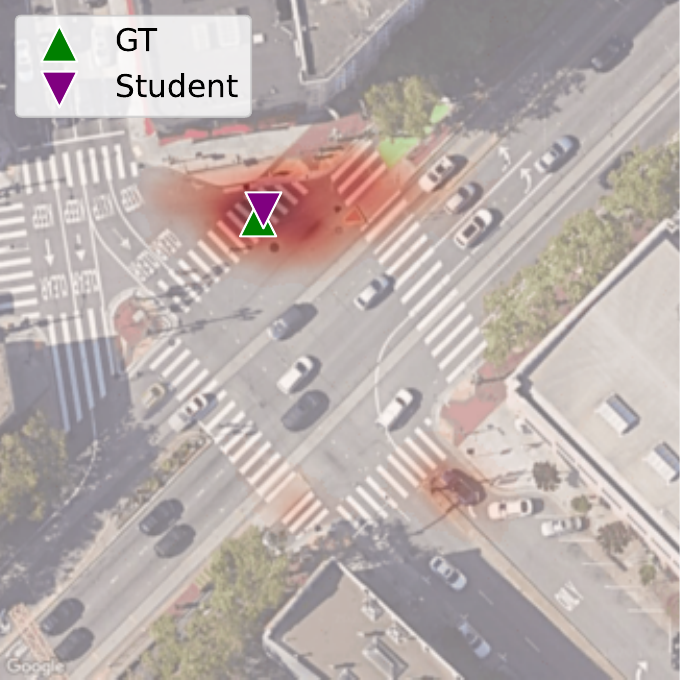}
    \includegraphics[align=c, width=0.24\linewidth]{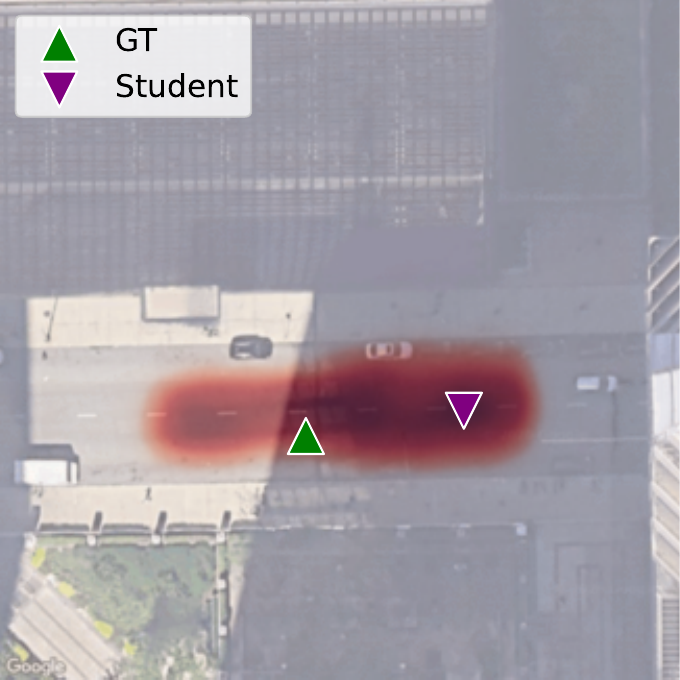}
    \end{minipage}
    \caption{Teacher and student models' predictions on VIGOR test set. The red color denotes the localization probability (a darker color means a higher probability). First three: success cases. Last: a failure case.}
    \label{fig:CCVPE_qualitative_results2}
\end{figure*}

Then, we visualize more teacher and student models' predictions in Figure~\ref{fig:CCVPE_qualitative_results2}.
The first two examples show a situation in which the teacher model's prediction contains multi-modal uncertainty, and the predicted location is in the wrong mode.
After weakly-supervised knowledge self-distillation, our student model assigns a higher probability at the correct mode.
In the third example, the teacher's prediction is accurate, and the student model maintains this accurate prediction.
Lastly, we showcase a challenging scenario where there lack of discriminative features, \eg the buildings in the aerial view mostly contain repetitive patterns. 
Although the teacher model picks a location close to the ground truth and the student has a higher error in this example, the inherent uncertainty in both the teacher's and student's heat maps is large.
We expect this can be addressed by using a sequence of ground-level images, and we will explore this in future work.


\section*{G. T-SNE Feature}
\begin{figure}[ht]    
\centering
    \includegraphics[width=6cm]{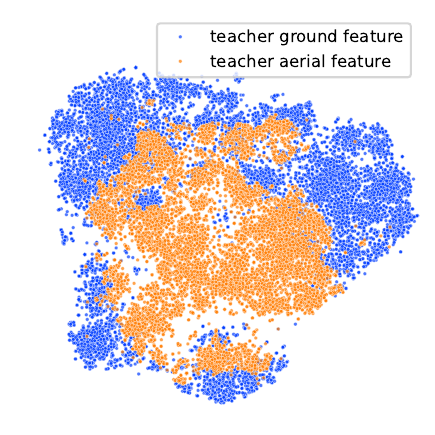}
    \includegraphics[width=6cm]{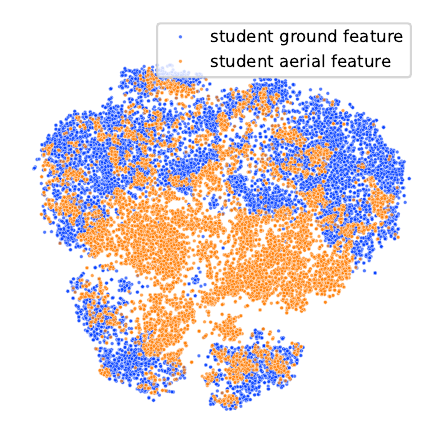}
    \caption{t-SNE, VIGOR test set: CCVPE teacher model (left) and final student model(right).}
    \label{fig:t-SNE}
\end{figure}

\reviewersugg{
To study if the extracted features by the teacher and final student models differ, we use t-SNE~\cite{van2008visualizing} to map the features to a two-dimensional space for visualization.
We collected CCVPE's ground features and the aerial features at the GT locations at the model bottleneck.
Figure~\ref{fig:t-SNE} shows their t-SNE plots before (teacher model) and after adaptation (final student model).
For the teacher model, ground and aerial samples are disjoint in the feature space, complicating matching across views.
For our student the plot shows more overlap between the two views, indicating better alignment.
This result supports that the quantitative improvement of our approach
results from adaptation to the target domain.
}

\section*{H. Potential Negative Impact}
Our paper proposed a weakly-supervised learning technique that enhances the localization accuracy of pre-trained fine-grained cross-view localization models.
Fine-grained cross-view localization techniques raise the risk of exposing precise location information of individuals.
For instance, mobile phone images, such as those from iPhones, often include a GNSS geo-tag in their metadata. This approximate location can be utilized to identify a local aerial image patch, thereby allowing fine-grained cross-view localization to pinpoint the exact location where the image was captured. 
Consequently, hackers could exploit this method to track individuals, such as social media influencers, by accessing the images they share online. 
This presents security and privacy concerns. 
To counter these risks, social media platforms should alert users to the potential for location data leakage and provide features that enable the removal of geo-tags from images upon upload.

\section*{I. Limitations}
In knowledge self-distillation, it is often required that the initial model is at a ``good enough'' starting point, otherwise, it will not converge to a better solution.
This requirement also applies to the method we propose. 
We conducted experiments where a teacher model, trained on one dataset such as KITTI~\cite{Geiger2013IJRR}, was used to generate pseudo ground truth to train a student model on a different dataset, for instance, the Ford dataset~\cite{agarwal2020ford}.
In this case, the teacher's predictions on the target dataset were not much better than random guesses, making our method not applicable. 
When the training and test sets are from different datasets, the teacher fails in the target area since the domain gap comes not only from different areas, but also from different sensors, and different resolutions of aerial images.
In our work, we target the domain gap between different areas but for the same sensor setup.

%
%
\bibliographystyle{splncs04}
\bibliography{main}
\end{document}